\documentclass[lettersize,journal]{IEEEtran}

\usepackage{textcomp}
\usepackage{stfloats}
\usepackage{verbatim}

\usepackage{tabularx,booktabs}
\usepackage{color}
\usepackage{tabularray}
\UseTblrLibrary{diagbox}
\usepackage{ifpdf}
\ifpdf
\else
\fi

\usepackage{cite}

\ifCLASSINFOpdf
\usepackage[pdftex]{graphicx}
\else
\fi

\usepackage{amsmath}
\usepackage{amsfonts}
\usepackage{amssymb}

\makeatletter
\newif\if@restonecol
\makeatother

\usepackage[linesnumbered,ruled,lined]{algorithm2e}

\usepackage{array}
\usepackage{multirow}
\usepackage{makecell,rotating}       
\usepackage{float}

\ifCLASSOPTIONcompsoc
\usepackage[caption=false,font=normalsize,labelfont=sf,textfont=sf]{subfig}
\else
\usepackage[caption=false,font=footnotesize]{subfig}
\fi

\usepackage{url}

\makeatletter
\let\NAT@parse\undefined
\makeatother
\usepackage{hyperref}  

\newtheorem{myDef}{Definition}

\begin{document}

\title{Differential Convolutional Fuzzy Time Series Forecasting}

\author{
	Tianxiang Zhan,
	Yuanpeng He, 
	Yong Deng,
	Zhen Li
	
	\thanks{Tianxiang Zhan:  Institute of Fundamental and	Frontier Science, University of Electronic	Science and Technology of China, 610054, Chengdu, China.}
	\thanks{Yuanpeng He:  Institute of Fundamental and	Frontier Science, University of Electronic	Science and Technology of China, 610054, Chengdu, China; Huawei Cloud, 523520, Dongguan, China.}
	\thanks{Zhen Li: China Mobile Information Technology Center, 100029, Beijing, China}
	\thanks{Corresponding author: Yong Deng,  Institute of Fundamental and	Frontier Science, University of Electronic	Science and Technology of China, 610054, Chengdu, China; School of Medicine, Vanderbilt University, Nashville,  37240, Tennessee, USA. Email address:  dengentropy@uestc.edu.cn and	prof.deng@hotmail.com}
}


\markboth{IEEE TRANSACTIONS ON FUZZY SYSTEMS}%
{Tianxiang Zhan, Yuanpeng He, Yong Deng \MakeLowercase{\textit{et al.}}: Differential Convolutional Fuzzy Time Series Forecasting}


\maketitle

\begin{abstract}
Fuzzy time series forecasting (FTSF) is a typical forecasting method with wide application. Traditional FTSF is regarded as an expert system which leads to loss of the ability to recognize undefined features. The mentioned is the main reason for poor forecasting with FTSF. To solve the problem, the proposed model Differential Fuzzy Convolutional Neural Network (DFCNN) utilizes a convolution neural network to re-implement FTSF with learnable ability. DFCNN is capable of recognizing potential information and improving forecasting accuracy. Thanks to the learnable ability of the neural network, the length of fuzzy rules established in FTSF is expended to an arbitrary length that the expert is not able to handle by the expert system. At the same time, FTSF usually cannot achieve satisfactory performance of non-stationary time series due to the trend of non-stationary time series. The trend of non-stationary time series causes the fuzzy set established by FTSF to be invalid and causes the forecasting to fail. DFCNN utilizes the Difference algorithm to weaken the non-stationary of time series so that DFCNN can forecast the non-stationary time series with a low error that FTSF cannot forecast in satisfactory performance. After the mass of experiments, DFCNN has an excellent prediction effect, which is ahead of the existing FTSF and common time series forecasting algorithms. Finally, DFCNN provides further ideas for improving FTSF and holds continued research value.

\end{abstract}


\section{Introduction}

\IEEEPARstart{T}{ime} series is a common data type, with univariable and multivariable sub-types. There are also many research directions of time series, such as time series clustering \cite{blazquez2021water, wang2022time, hayashi2022ocstn}, anomaly detection \cite{liu2022time, karczmarek2022choquet}, time series entropy \cite{peng2022characterizing, cui2022belief}, time series forecasting \cite{hernandez2020forecasting, wang2022trend, wang2022time} and so on \cite{zhang2022interpretable}. Time series forecasting has attracted the attention of many scholars and experts. Time series forecasting is useful in a wide range of fields, such as stock forecasting \cite{huang2021natural}, wind forecasting, population forecasting and so on. At present, there are many forecasting methods, such as the common statistical methods including exponential smoothing (ETS) model \cite{billah2006exponential} and Holt-Winter ETS model \cite{jiang2020holt}, autoregression (AR) \cite{xu2019modeling},  moving average model (ARIMA) \cite{domingos2019intelligent} and seasonal ARIMA model \cite{liu2021short}. However, the forecasting accuracy of the common model is not enough to meet needs currently. In order to improve the accuracy of forecasting, increasingly novel and advanced methods have been developed, such as the method based on machine learning \cite{masini2021machine}, neural network \cite{huang2021new} and complex network. How to further improve the accuracy of time series forecasting is an open problem. 

Fuzzy time series forecasting (FTSF)  is  a far-reaching time series method based on fuzzy theory. Fuzzy theory has the ability to deal with the uncertainty of data. It is widely used in engineering practice, such as game theory \cite{ponce2020tailored,wu2021strategies,xu2022game}, intelligent decision-making \cite{Xiao2021GIQ,albahri2022novel,xiao2020efmcdm,kumar2022multiple}, information fusion \cite{Xiao2022Acomplexweighted,Xiao2022GEJS}, data classification \cite{zhang2020boosting,zhang2021lfic,xiao2021distance}, bioinformatics \cite{rashid2022novel, barua2023automated}, pattern classification \cite{Xiao2022NQMF,Xiao2022Generalizeddivergence} and so on \cite{cheng2022ranking, lin2022picture}.  The previous FTSF works have a common fact that the process of fuzzifying a  variable and the process of defuzzifying a variable must be certain. Furthermore, there is only reasoning between fuzzy variables which leads to a high possibility of prediction bias due to fuzzy set creation. In addition, due to reasoning among the mass of fuzzy rules, reasoning as a expert system is not an optimal way that there are potential differences and inherent relations. Besides, the length of fuzzy reasoning is also limited by the mode of the expert system, since the expert system is not able to learn pattern of longer fuzzy rules. Therefore, it is proper to utilize optimization algorithm to further reasoning. One of the implementations of the optimization algorithm is the neural network.

Stationary is a crucial feature of time series, and time series are usually non-stationary due to trend and so on. According to the method of Cleveland et al. \cite{cleveland1990stl}, time series can be decomposed into residuals component, seasonal component and trend component. Because of the trend component, the tendency of a time series to change in one direction over time. As mentioned above, FTSF can only reason about fuzzy variables. When the time is far from the time when the fuzzy system is established, the variables will be fuzzified to first or last fuzzy set of the previously established fuzzy set because of the accumulated trend. Then, each time will be regarded as the same fuzzy variable and defuzzifying process leads to the same forecasting output which called fuzzy failure (FF). To solve FF problem, on the one hand, the time series need to remove the trend, and on the other hand the expert system approach needs to be replaced, because expert systems lacks knowledge summarization and autonomous learning capabilities.

To solve the mentioned problem, Differential Fuzzy Convolutional Neural Network (DFCNN) is proposed. In DFCNN, the data are differenced, which is an effective method to remove time series trends. To solve the problem that the expert system can only reason about fuzzy variables, the DFCNN utilizes the neural network to optimize the FTS. Recently, some works proposed FTSF with Artificial Neutral Network (ANN) \cite{rahman2015artificial, singh2018rainfall, jiang2019hybrid, kocak2020new, bas2022novel}. FTSF with ANN is limited by the optimization object of ANN fuzzy variables \cite{egrioglu_new_2009,aladag2009forecasting, aladag2010high, chen2011multivariate, egrioglu2013fuzzy}. Compared with the combined FTS and ANN approach, the proposed method skips the establishment of fuzzy sets, making the inference not limited to fuzzy variables. Proposed Fuzzy Generator with reference to the fuzzy treatment of traditional FTS prediction methods. Fuzzy Generator divides the fuzzy interval with the construction of coding theory, which makes the input time series expand into Fuzzy Token. Fuzzy Token is a type of vector remained the full information of converting time series to fuzzy variables. The convolutional layer is a mechanism used to replace the membership functions while making the inference not limited to fuzzy variables. For the problem of the length of fuzzy inference, Sliding Window algorithm is applied in DFCNN to process the data, and the window size is equivalent to the length of the fuzzy rules of traditional FTS. By processing the convolutional layer, DFCNN can handle fuzzy rules that are not restricted to the previous binary fuzzy rules and can build longer fuzzy inference. So the contribution of this work is as follows:
\begin{itemize}
	\item Proposing Fuzzy Generator solves the limitation of fuzzifying variables.
	\item Utilizing the Convolution layer as membership functions solves the limitation of expert system.
	\item Utilizing the Sliding Windows algorithm expands the limitation of length of fuzzy rules.
\end{itemize}

In the experimental section, many FTSF methods are selected to be compared methods. To avoid the effect of trend, each compared method was appended the Difference algorithm. In terms of prediction error, DFCNN exceeds the comparison method.  DFCNN is also compared with common time series forecasting methods. In the vast majority of cases, the predictive performance of DFCNN is stable. However, in one of the small datasets DFCNN is not able to achieve the most accurate prediction. The limitations of DFCNN are analyzed in this dataset, as well as for the prediction errors of the comparison methods. Meanwhile, the effects of the number of convolutional kernels and window size on DFCNN predictions were experimented. The window size directly affects the prediction accuracy and does not show a significant correlation with the selected data and correlation, and the prediction is relatively stable under the default parameters. In a word, the experimental results demonstrate that the comprehensive prediction error of DFCNN is low, which is better than the comparison methods.

The  structure of the article is as follows. In the second section, it explains the previous methods and problems of fuzzy time series forecasting. In the third section, the basic theory of the proposed method is introduced. In the fourth section, the article proposes DFCNN. In the fifth section, the article shows the experimental effect of DFCNN and the comparison and analysis with traditional fuzzy time series forecasting. In the sixth section, the article summarizes the full text and looks forward to future work. 

\section{Previous Fuzzy Time forecasting Research and Problems}

Fuzzy time series forecasting is a common method in time forecasting. The initial fuzzy time series was used for enrollment forecasting of  University of Alabama \cite{chen1996forecasting}. The data size of the previous experiments was small, and in this case, the pure use of fuzzy logic can effectively predict future enrollment. In today's era of mega data, it is possible to meet FF problems when use FTSF algorithm.

\begin{figure}[htbp]
	\centering
	\resizebox{0.85\linewidth}{!}{
		\includegraphics{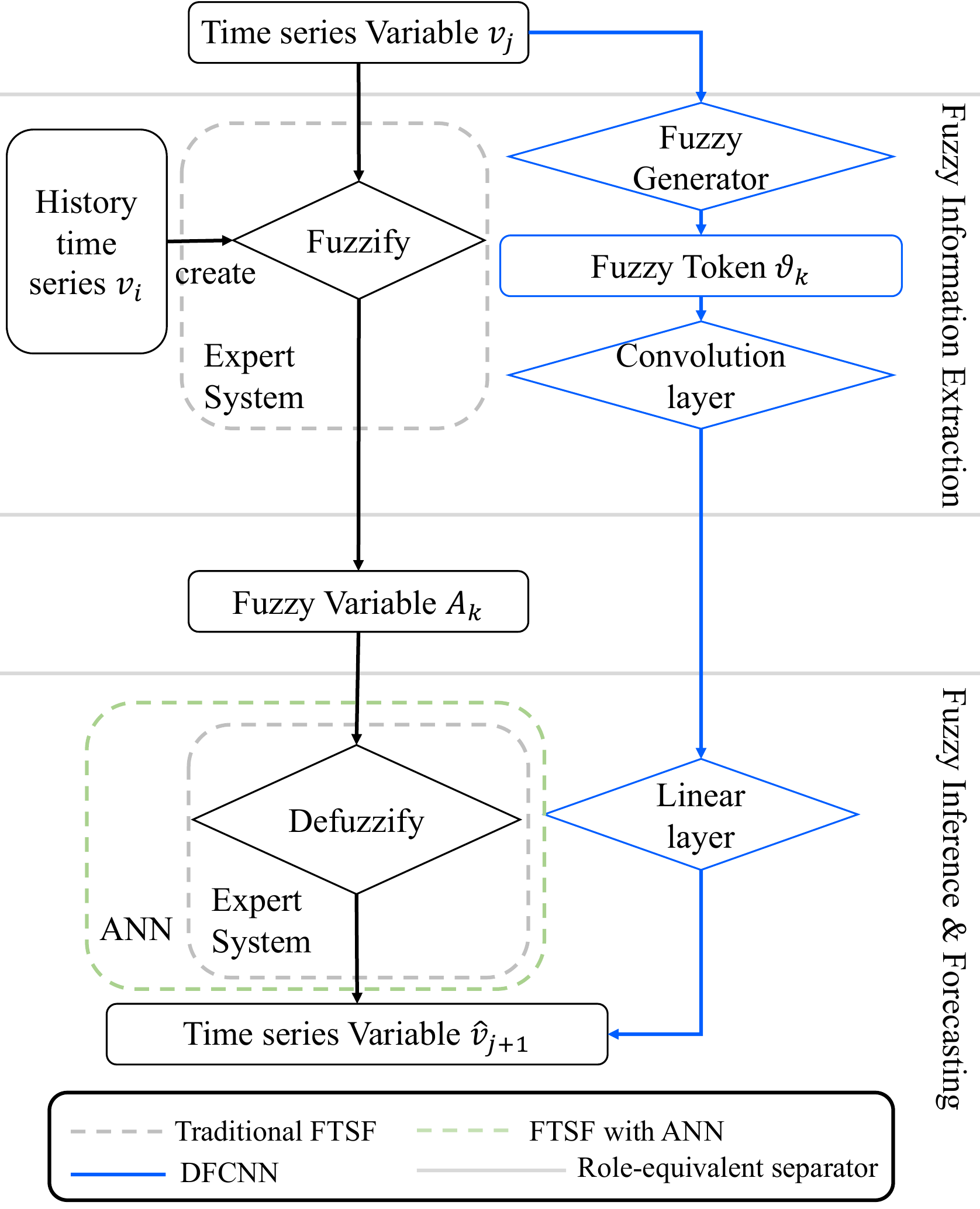}}
	\caption{Flow chart of FSTF and DFCNN}
\end{figure}

\subsection{Process of fuzzy time series forecasting}
First assume that the time series is T, and T is defined in Eq.1:
\begin{equation}
	T = \left\{(t_1,v_1),(t_2,v_2),(t_3,v_3),...,(t_n, v_n)\right\}
\end{equation}
where $t_i$ is the time of time node $i$, $v_i$ is the value of time node $i$ and $n$ is the size of time series $T$.

Taking Chen's methods \cite{chen1996forecasting} as an example, the fuzzy time series forecasting method usually adopts the following steps:
\begin{enumerate}
	\item An interval $U$ of a time series is defined so that the maximum values $MAX(T)$ and minimum values $MIN(T)$ of the time series belong to this interval as Eq.2. $\sigma$ is an arbitrary positive number.
	\begin{equation}
	U =	\left[MIN(T) -\sigma, MAX(T)+\sigma\right] 
	\end{equation}
	
	\item According to the proposed method, the interval $U$ is divided into some sub-intervals $U_1, U_2, U_3,...,U_k$.  The number of divisions $k$ is an arbitrary positive number.
	
	\item Establish fuzzy sets $A_1, A_2, A_3,...,A_k$. The establishment of the fuzzy set and the interval established in the second step are related by the membership function.
	
	\item Fuzzify time series into fuzzy linguistic variable series $A_i, A_j,...A_k$.
	
	\item The inference system is built by forming fuzzy rules $A_i \rightarrow A_j,...,A_m \rightarrow A_k$ from two consecutive linguistic variables  in  the fuzzy linguistic variable series.
	
	\item Assume current time is $t_n$. Fuzzy current time value $v_n$ into corresponding fuzzy linguistic variable $A_k$. According the existed fuzzy rules, inference system generates a certain fuzzy rule $A_k \rightarrow A_m$. 
	
	\item Defuzzify $A_m$ into $\hat v_{n+1}$ and output $(t_{n+1}, \hat v_{n+1})$.

\end{enumerate}

\subsection{Existing problems}

\subsubsection{Limitation of fuzzy reasoning}
According to the forecasting process of fuzzy time series, the input of fuzzy time series forecasting is the current time node $(t,v)$ and the fuzzy set $A$ to which the current node belongs.  The fuzzy set $A$ can be defined by membership functions and intervals $u$. The membership function belongs to the process of fuzzy time series forecasting, and the input of the fuzzy time series forecasting is the time node $(t,v)$ and the interval $u$ to which the time point belongs. 

The forecasting method of fuzzy time series can be transformed as a function  "Fuzzy Time Series Forecasting" ($FTSF$) in Eq.3, Eq.4.
\begin{equation}
	\hat v = FTSF(t,v,u)
\end{equation}

\begin{equation}
	u = [b_{l}, b_{r}]
\end{equation}
where $b_{l}, b_{r}$ are the left and right boundaries of interval $u$.

The flow chart of FTSF is left side of Fig.1. It means that the fuzzy inference logic of FTSF is independent of the original data directly. It is also a limitation of expert system which is unable to learning data feature by itself. Also there are some FTSF methods based on ANN, and fuzzy inference is improved. But FTSF methods based on ANN is not able to solve the improper process when fuzzy sets is establishing.

\subsubsection{Fuzzy failure}
FF problem is mentioned. Different time series have different properties, such as periodicity and trend. Chen's method was an example of FTSF and forecast the Association of Securities Dealers Automated Quotations (NASDAQ) average \cite{chen1996forecasting} in Fig.2. There is a FF problem. Each time is regarded as the same fuzzy variable and the same defuzzifying process then output the same result. 

\begin{figure}[htbp]
	\centering
	\resizebox{0.8\linewidth}{!}{
		\includegraphics{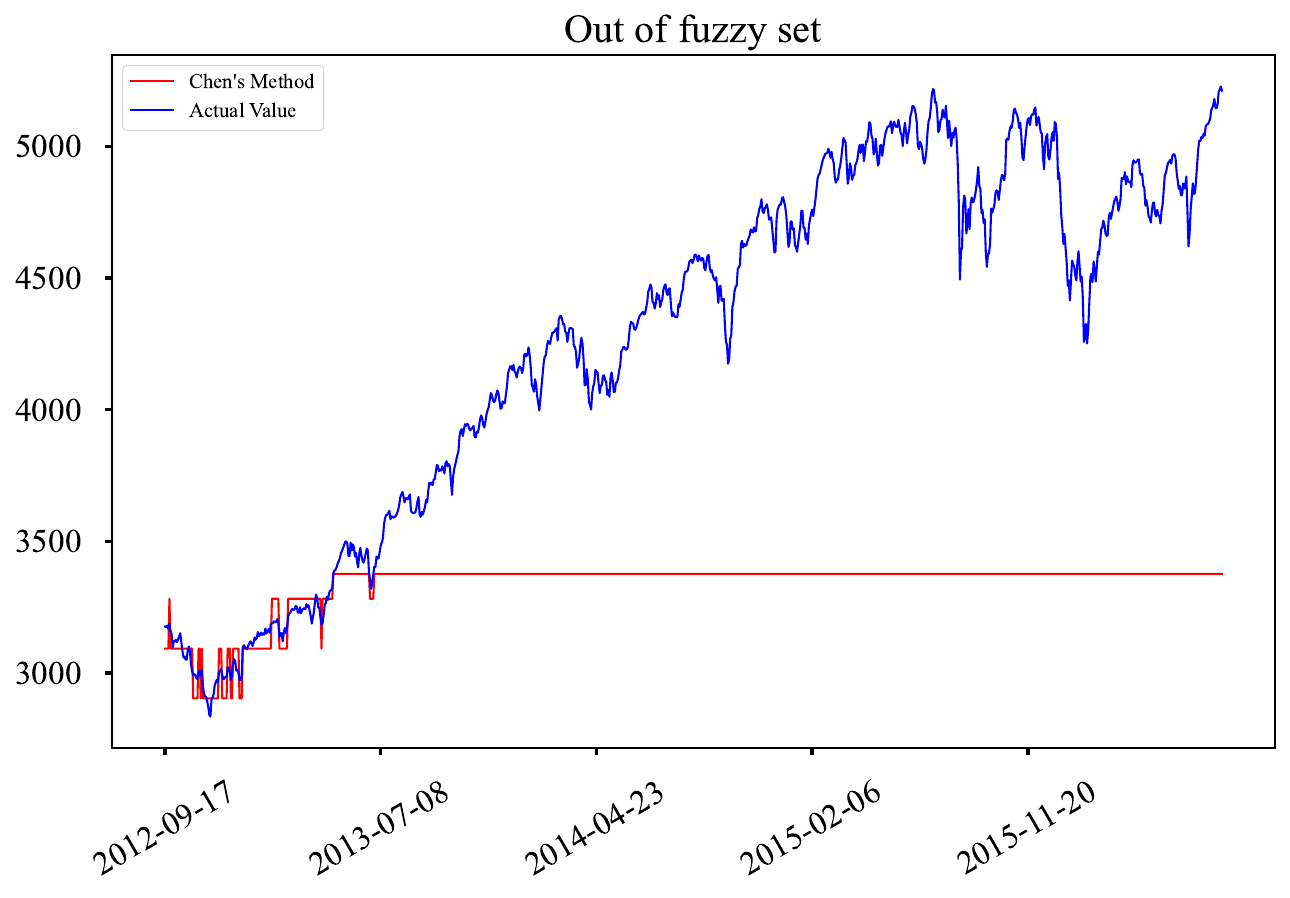}}
	\caption{The effect of the time node out of fuzzy set}
\end{figure}

\subsubsection{Limitation of length of fuzzy rules}

When FTSF method fit to an existed time series need to generate many fuzzy roles like $A_i \rightarrow A_j$. It is base of fuzzy inference and defuzzifying. But there is no reason not to consider how to process fuzzy rules which length is large than 2. Arbitrary positive number large than 2 is a possible length of fuzzy rules, but FTSF is an expert system method which means it need to proposed certain reasoning logics for each arbitrary positive number.

\section{Preliminary}

\subsection{Difference algorithm}
The difference algorithm takes the difference between each time node in the time series $T$ , while retaining the value of the first time node as Eq.5 .

\begin{equation}
	DIFF(T) = \left\{(t_2,v_2-v_1),...,(t_n,v_n-v_{n-1})\right\}
\end{equation}

The difference series effectively removes the trend of the time series, and the initial value of the time series is recorded. If you restore the original time series, it only need to add the difference subseries values before each node. 

%

\subsection{Convolution}

In deep learning, convolution operation is a common operation for feature mapping \cite{Goodfellow-et-al-2016}. Now suppose the input of the convolution operation is $I$ and the kernel function is $K$.  The convolution operation is usually denoted by the symbol $\ast$. The output of the convolution operation is denoted as $S$ in Eq6, Eq.7. 
\begin{equation}
	S = I \ast K
\end{equation}
\begin{equation}
	S(i,j) = (I \ast K)(i,j) = \sum_m \sum_n I(i+m, j+n)K(m,n)
\end{equation}

Where $S$, $I$, $K$ are in matrix form. The convolution operation will continuously update the weight of the kernel function $K$ after supervised learning and multiple cashback propagation, and finally output the features of the map. If $I$ is a vector, the output $S$ of the convolution is calculated according to Eq.8.

\begin{equation}
	S(i) = (I \ast K)(i) = \sum_m I(i+m)K(m)
\end{equation}


\section{Differential Fuzzy Convolutional Neural Network}

A time series $T = \left\{(t_1,v_1),(t_2,v_2),...,(t_n, v_n)\right\}$ is assumed.
\subsection{Difference Layer}

The sequence obtained after differencing the time series $T$ is called $D$ as Eq.5.

Difference Layer is capable of removing the trend and time series is more stationary. The first value $v_1$ of time series needs to reserve for recover the original time series from difference series.

\subsection{Sliding Window Layer}
The formula for sliding window algorithm is Eq.9, where $i$ refers to the i-th result of the sliding window and $x$ is the original input.

\begin{equation}
	SlidingWindow(i,lookback) = \left\{x_i, x_{i+1}, ..., x_{i+lookback-1} \right\}
\end{equation}

Suppose that the two adjacent values of the original time series are $v_1$ and $v_2$. The fuzzy variable corresponding to $v_1$ is $A$ and the fuzzy variable corresponding to $v_2$ is $B$. So the resulting fuzzy rule is $A \rightarrow B$. When $lookback=2$, Sliding Windows algorithm's first output is $(v_1,v_2)$ which is equivalent to defuzzified $A \rightarrow B$ .

Suppose the value of the original time series after $v_2$ is $v_3$. The fuzzy variable corresponding to $v_3$ is $C$. Then the fuzzy rule becomes $A \rightarrow B \rightarrow C$. When $lookback=3$, Sliding Windows algorithm's first output is $(v_1,v_2,v_3)$ which is equivalent to defuzzified $A \rightarrow B \rightarrow C$ .

There is no good way to deal with fuzzy rules which length is over 2 and make forecastings based on fuzzy rules. Sliding Windows algorithm slove the problem of limitation of fuzzy rules' length. But Silding Windows algorithm is not able to consider any information of original fuzzy sets. So Fuzzy Generator is a way to embed the necessary information of establishing fuzzy set instead of create fuzzy sets and fuzzy variables directly.

\subsection{Fuzzy Generator Layer}

\myDef{Encoding Partition}

For a given fuzzy interval $U=[L,R]$, assuming corresponding time series $T$ length is $n$, then the number of divided fuzzy interval is $\lceil log_2(n) \rceil$. The fuzzy intervals boundaries set $N$ is shown in Eq.10. The partitioned fuzzy intervals are all of equal length.

\begin{equation}
	\begin{split}
		N
		=&\left\{n_1,n_2,...,n_{\lceil log_2(n) \rceil+2}\right\}  \\
		=&\{ L, L + \frac{R-L}{\lceil log_2(n) \rceil+2},  ...,L + m*\frac{R-L}{\lceil log_2(n) \rceil+2},R \}
	\end{split}
\end{equation}

Here is the proof of this definition. Suppose the fuzzy set membership is either 1 or 0 in Eq.11.
\begin{equation}
	\mu(T_i) \rightarrow \left\{0, 1\right\}
\end{equation}
where $\mu$ is the membership function of arbitrary fuzzy set and $T_i = (t_i, v_i)$ is the i-th time node in time series $T$.

\begin{figure}[htbp]
	\centering
	\resizebox{0.8\linewidth}{!}{
		\includegraphics{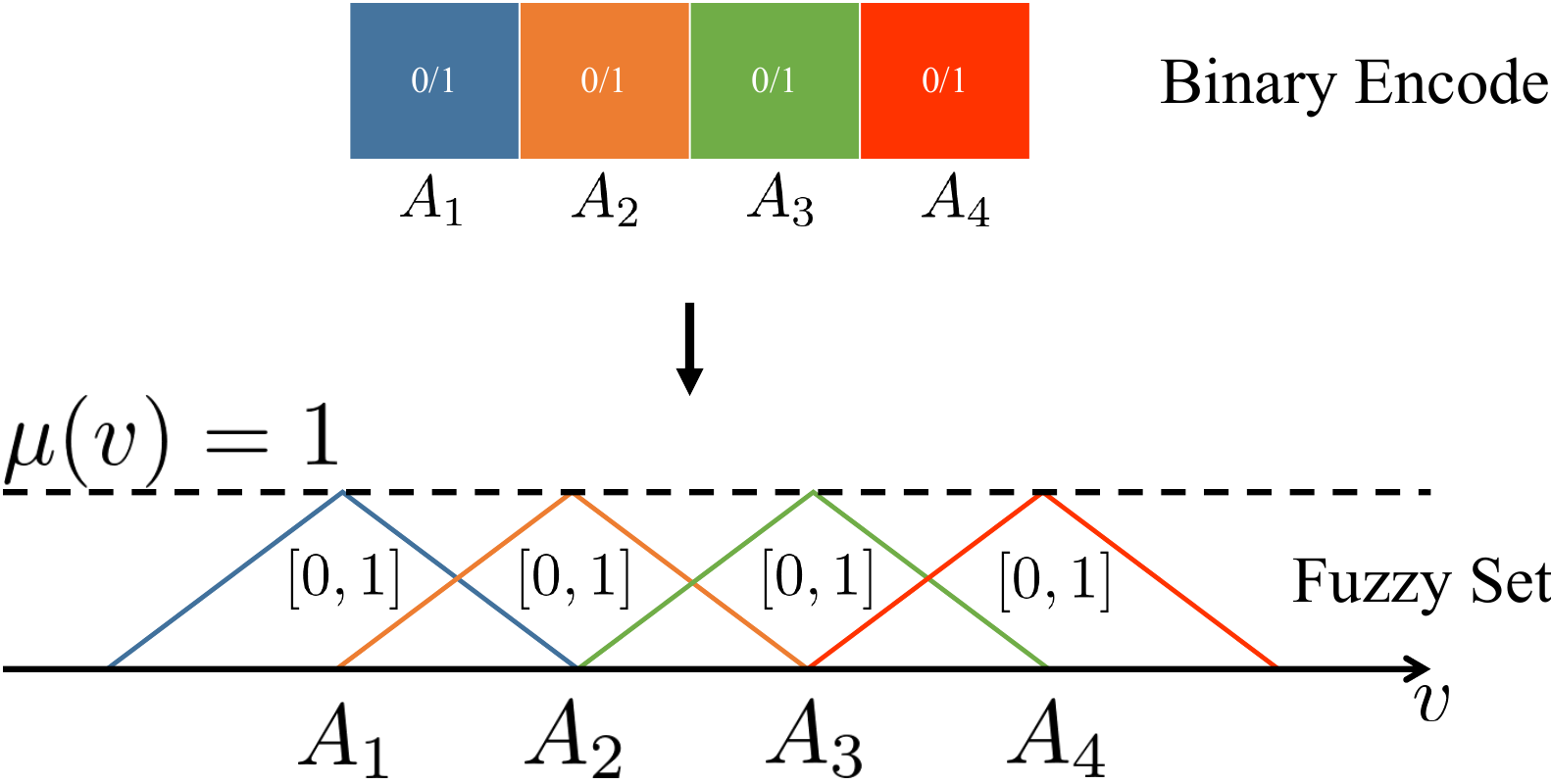}}
	\caption{The transformation from original time series to difference series}
\end{figure}

Then if fuzzy sets needs to be used to encode a time series $T$ of length $n$ bits, the required number $m$ of fuzzy sets is calculated in Eq.12, Eq.13, Eq.14:
\begin{equation}
	2^m \geq n
\end{equation}
\begin{equation}
	m \in Z
\end{equation}
\begin{equation}
	m = \lceil log_2(n) \rceil
\end{equation}

This method of calculating the number of fuzzy sets is the same as that of binary encoding \cite{song2022dynamic}.  However, the membership function of a fuzzy set is a continuous map. The original fuzzy set is used as a bit, which can represent more states. Fig.3 shows an example of partitioning fuzzy sets by coding ideas. Therefore, the number of fuzzy intervals is split according to $m = \lceil log_2(n) \rceil$, which is able to fully encode all the nodes in the time series.

\textbf{Example} Assume a time series T in Eq.15
\begin{equation}
	T = \left\{(1,2),(2,3),(3,5),(4,6),(5,4),(6,7)\right\}
\end{equation}
where $T$ is composed of $6$ time points $(t_i. v_i)$. The number of fuzzy intervals $m$ in Eq.16  required for the time series $T$ is
\begin{equation}
	m = \lceil log_2(6) \rceil = 3
\end{equation}
According to Eq.16, the resulting fuzzy set is assumed to be three equally long fuzzy intervals $A_1, A_2, A_3$.

\begin{equation}
	L=MIN(D)-STD(D)
\end{equation}

\begin{equation}
	R=MAX(D)+STD(D)
\end{equation}

\begin{equation}
	U = [L,R]
\end{equation}

Domain $U$ is defined in Eq.17, Eq.18, Eq.19 which $MIN(D)$ refer to the minimum value $v_{min}$ in time series values, also $MAX(D)$ refer to the maximum value $v_{max}$ and $STD(D)$ refer to standard deviation. The quantity of \textbf{partial coincident} fuzzy intervals is $m$ in Eq.20 that  the length of the time series is reduced to $(n-1) $ due to Differential Layer.  The interval $U$ is then divided into $(m+1)$ \textbf{uncoincident} segments of equal length. There are $(m+2)$ boundaries for these equally long intervals, and the set of boundaries is denoted by $N$ in Eq.21 as Fig.4.

\begin{equation}
	m = \lceil log_2(n-1) \rceil
\end{equation}

\begin{equation}
	\begin{split}
		N
		=&\left\{n_1,n_2,...,n_{m+2}\right\}  \\
		=&\{ L, L + \frac{R-L}{m+1},  ...,L + m*\frac{R-L}{m+1},R \}
	\end{split}
\end{equation}

\begin{figure}[htbp]
	\centering
	\resizebox{\linewidth}{!}{
		\includegraphics{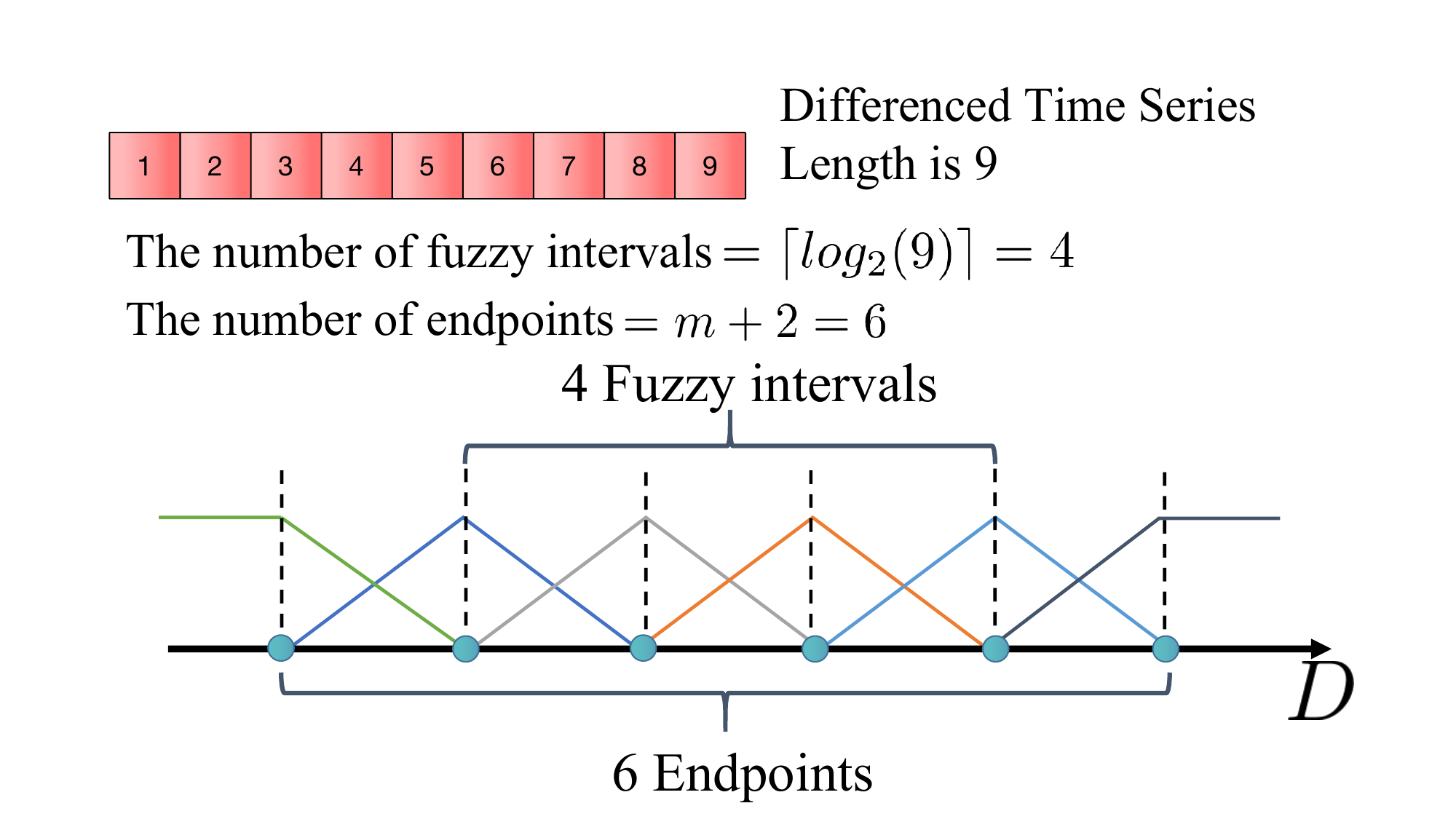}}
	\caption{The establishment process of fuzzy interval for a difference series of length 9. The triangle is shown as fuzzy membership $\mu(x)$, which is \textbf{skipped} in the DFCNN. Here it is only considered that the output of each membership function determined by the input $x$ depends on the two endpoints where $x$ is nearest.}
\end{figure}

\myDef{Fuzzy Token}

Fuzzy Token is a vector contains the information of establishing fuzzy set and fuzzy variable. The nearest fuzzy intervals boundaries of time series values are embedded into Fuzzy Token. The formula of Fuzzy Token is Eq.22, Eq.23 and Eq.24.

\begin{equation}
	FuzzyToken(v) = (l,v,r)
\end{equation}

\begin{equation}
	l = \mathop{\arg\min}\limits_{x} (\vert x-v  \vert = \underset{y \in N\ and\ y<v}{{\min} \vert y-v  \vert})
\end{equation}

\begin{equation}
	r = \mathop{\arg\min}\limits_{x} ( \vert x-v  \vert = \underset{y \in N\ and\ y>v}{{\min} \vert y-v  \vert})
\end{equation}

Time series value $v$ is converted into the Fuzzy Token $(l, v, r)$ which $l,r$ are the two nearest boundaries in fuzzy intervals boundary set $N$ mentioned by Fuzzy Generator. The definition of $l$ and $r$ is shown in Eq.23, Eq.24. $l$ and $r$ has two cases to be defined shown in Fig.5.  If time series value $v$ is in the boundary set $N$, for $v$, generating Fuzzy Token ignores $v$ in $N$ and also find the nearest two boundaries.

\begin{figure}[htbp]
	\centering
	\resizebox{0.8\linewidth}{!}{
		\includegraphics{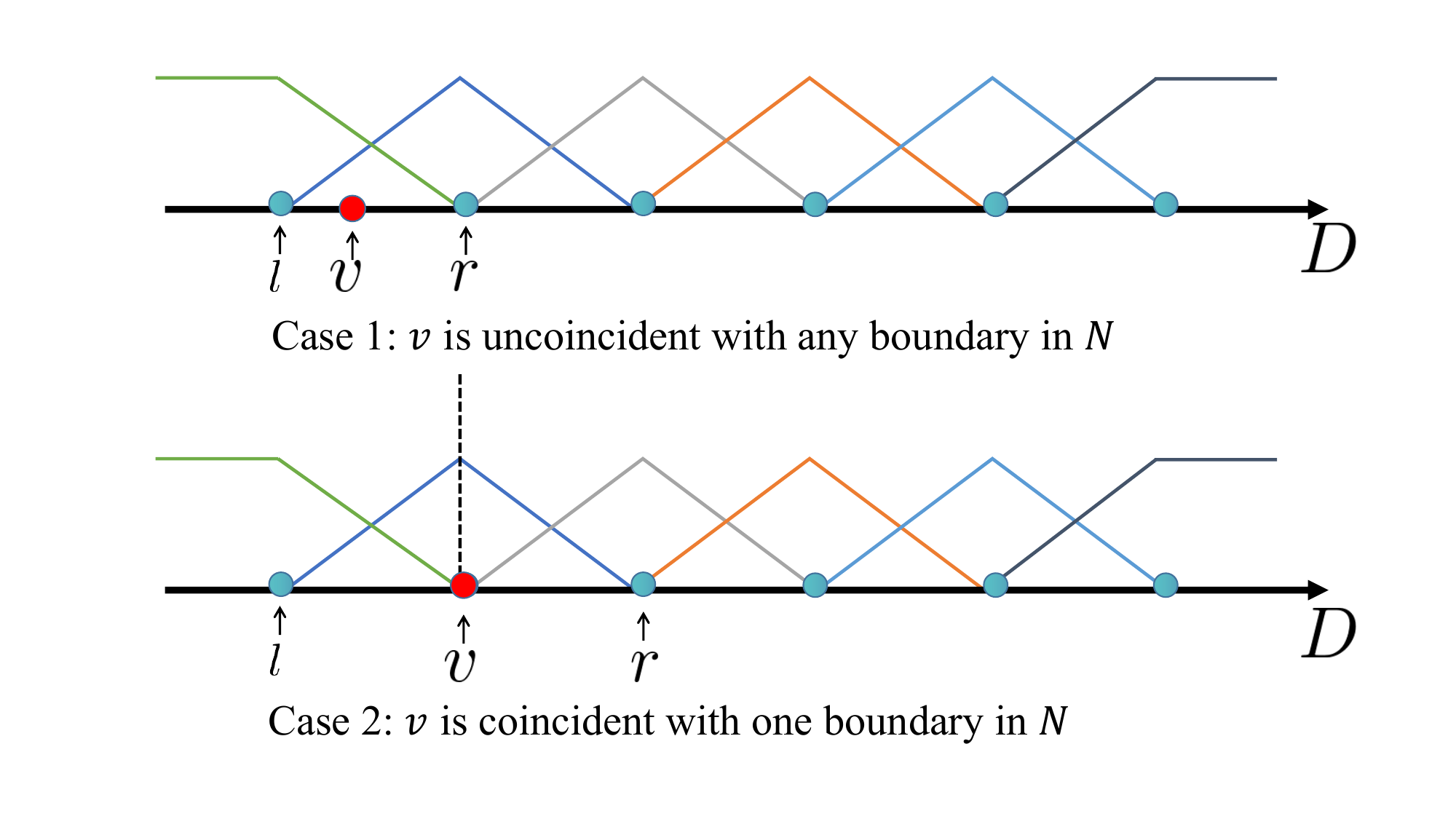}}
	\caption{There are two cases of Fuzzy Token of time values $v$.  The only difference of two cases is whether $v$ is in the fuzzy interval boundaries set $N$. In other words, if time series value $v$ is coincident with one boundaries in $N$, the boundary $v$ is ignored in $N$ for $FuzzyToken(v)$.}
\end{figure}


\begin{algorithm}[htbp] 
	\caption{Differential Fuzzy Convolution Neural Network}
	\KwIn{Time series $T$, lookback $l$}
	\KwOut{Forecasting value $\hat y$}
	\tcp{Difference time series}
	$D = DIFF(T)$\;
	\tcp{Sliding window processing}
	$D_{window} = SlidingWindows(T,l)$\;
	\tcp{Computed left and right boundary}
	$l=[min(D)-std(D)]$ \;
	$r=[max(D)+std(D)]$\;
	\tcp{Calculate the number of fuzzy intervals}
	$n = length(D)$\;
	$m = \lceil log_2(n) \rceil$\;
	\tcp{Partition fuzzy interval}
	$N = linspace(low,up,m+1)$\;
	\tcp{Initializes the fuzzy token set}
	$F_t = \{\}$\;
	\tcp{Create fuzzy tokens}
	\For {Each Window $D_{window,i}$ }
	{
		\tcp{Initializes a temporary fuzzy token collection in a window}
		$F_{window,t} = \{\}$\;
		\tcp {Handles fuzzy tokens in the window}
		\For {Each Value in the Window $v_j$ }
		{
			$l =  \underset{x \in N\ and\ x<v}{{\arg\min} \, \vert x  \vert} $\;
			$r =  \underset{x \in N\ and\ x>v}{{\arg\min} \, \vert x  \vert} $\;
			$F_{window,t} = \{l, v_j, r\}$\;
		}
		$F_t(i) = F_{window,t}$\;
	}
	\tcp{Normalize the fuzzy token}
	$F_t' = Norm(F_t)$\;
	\tcp{Fuzzy rules for convolution learning}
	$F_t'' = Conv(F_t')$\;
	\tcp{Synthesis and forecasting}
	$\hat y = sum(F_t'')A^T+b$\;
	return $\hat y$ \;
\end{algorithm}

\subsection{Batch Norm Layer}
For the Fuzzy Token $x$, it is a optimal to normalize $x$ as Eq.26 \cite{ioffe2015batch, awais2020revisiting} which is able to accelerate the process of learning process. The mean $E(\cdot)$ and standard-deviation  $Var(\cdot)$ are calculated. $\gamma$ and $\beta$ are learnable parameter vectors of size which is the number of features or channels of the input. The normalized Fuzzy Token is written as $x'$.
\begin{equation}
	x' = Norm(x) = \frac{x-E(x)}{\sqrt{Var(x)+\epsilon}}*\gamma+\beta
\end{equation}

\subsection{Convolution Layer}

Fuzzy Generator combines the value of traditional fuzzy variable, and convolution operation can be used as a type of fuzzy function. Convolution Layer is a one-dimensional Convolution Layer which is equivalent to membership function with learning parameter weight $w$ and bias $b$.

\begin{equation}
	\mu_1(x) = \left\{\begin{matrix}
		&&1\ \ && x \leqslant n_1\\
		&&\frac{n_2-x}{n_2-n_1}\ \ &&n_1<x \leqslant n_2
	\end{matrix}\right.
\end{equation}

\begin{equation}
	\mu_2(x) = \left\{\begin{matrix}
		&&\frac{x-n_1}{n_2-n_1}\ \ &&n_1<x \leqslant n_2  \\
		&&\frac{n_3-x}{n_3-n_2}\ \ &&n_2<x \leqslant n_3
	\end{matrix}\right.
\end{equation}

\begin{equation}
	\mu_3(x) = \left\{\begin{matrix}
		&&\frac{x-n_2}{n_3-n_2}\ \ &&n_2<x \leqslant n_3 \\
		&&1\ \ && x > n_3
	\end{matrix}\right.
\end{equation}

Fuzzy variable is determined membership function which is determined by fuzzy interval and membership function in FTSF. Eq.26, Eq.27 and Eq.28 is common membership functions of fuzzy sets. Assume $n1, n2, n3$ is the boundaries of corresponding fuzzy set $A_1, A_2, A_3$. For a  fuzzy intervals like $[n1,n2)$, the membership function $\mu_1(x)$ of $A_1$ is Eq.29. 

\begin{equation}
	 \begin{split}
		(A_1, \mu_1(x))&=(A_1, \frac{n_2-x}{n_2-n_1}) \\
		&=(A_1, -\frac{1}{n_2-n_1}x + \frac{1}{n_2-n_1}*n_2 ) \\
		&\Leftrightarrow (A_1,w_1*x+w_2*r) \\
		&\Leftrightarrow (A_1,\sum (w_i*FuzzyToken(x) +b_i)) \\
		&\Leftrightarrow (A_1,Conv(FuzzyToken(x)))
	\end{split}
\end{equation}

$\frac{1}{n_2-n_1}$ is a constant variable. To optimal fit to input time series, $\frac{1}{n_2-n_1}$ needs to be treat as a learnable parameter $w$. $n_1, n_2$ is same as the left and right boundaries $r, l $ in Fuzzy Token. So one dimension Convolution Layer is equivalent mechanism of fuzzy variables.

Because of the sliding window, the length of the input for the convolution layer is $batch*lookback*3$, where the number of channels is $lookback$ and feature size is 3.  The convolution layer has a parameters, the number of output channels of the convolution  $\theta$, which is also the number of convolution kernels. The number of convolution kernels $\theta$ can be set to multiple and is usually set to 2. Multiple convolution kernels can increase the robustness of the network. Convolution kernel size is $1*3$, and convolution kernel sums the all channel which gathers information of fuzzy rules in Eq.30. And  the length of equivalent fuzzy rule is not limited.

\begin{equation}
		 \begin{split}
		 	& A_i \rightarrow A_j  \rightarrow ...\\
		 	& \Leftrightarrow Conv(FuzzyToken(x_i)) \rightarrow Conv(FuzzyToken(x_j)) \\
		 	& \rightarrow...\\
		 	& \Leftrightarrow \sum Conv(FuzzyToken(x_i))\\
		 	& \Leftrightarrow Conv(FuzzyToken(x))
		 \end{split}
\end{equation}

In right side of Fig.1, Fuzzy Generator and Convolution layer is separated in fuzzy information extraction block. It is different to FTSF's Fuzzifying process based on expert system for Fuzzy Generator and Convolution layer that the relevant learnable parameter helps fit the input time series further.

\subsection{Linear Layer}

To predict the next time point, in FTSF,  the fuzzy rules should be weighted according to the existing fuzzy rules and output. The output of the convolution is summed and the final forecasting is made through the linear layer. Assuming the output of the convolution layer is $x ''$, the last dimension needs to added weight and sum up. Similar to FTSF, linear layer adopts a weighted way to predict the previously created fuzzy rules in Eq.31. The overall structure of the network is shown in Fig.6.

\begin{equation}
	\hat{y'} = sum(x'')A^T+b
\end{equation}

In left side of Fig.1, in the fuzzy inference and forecasting block, traditional FTSF's defuzzifying process is based on expert system. Also FTS with ANN utilize ANN to output the proper label of target fuzzy set and use same method as traditional FTSF to defuzzyify.  In essence,  FTS with ANN does not have enough learning ability.

\subsection{Recovery}
The target of Linear Layer is difference series $y'_{t+1}$. The recovery process is necessary for output in Eq.32.

\begin{equation}
\hat y_{t+1} =  Recovery(\hat  y'_{t+1}) = \hat  y'_{t+1} +  y_t
\end{equation}

$y_t$ is origin time series of previous one time step of $y'$. In the same way as the FTSF, DFCNN is a single step forecasting method due to Recovery mechanism. To keep stationary, Difference algorithm is applied in DFCNN also leads to single step forecasting. If Difference algorithm is replaced by other algorithm which transforms time series to stationary time series and does not rely on one time step values, the multistep forecasting can be implemented.

\begin{figure*}[htbp]
	\centering
	\resizebox{0.8\linewidth}{!}{
		\includegraphics{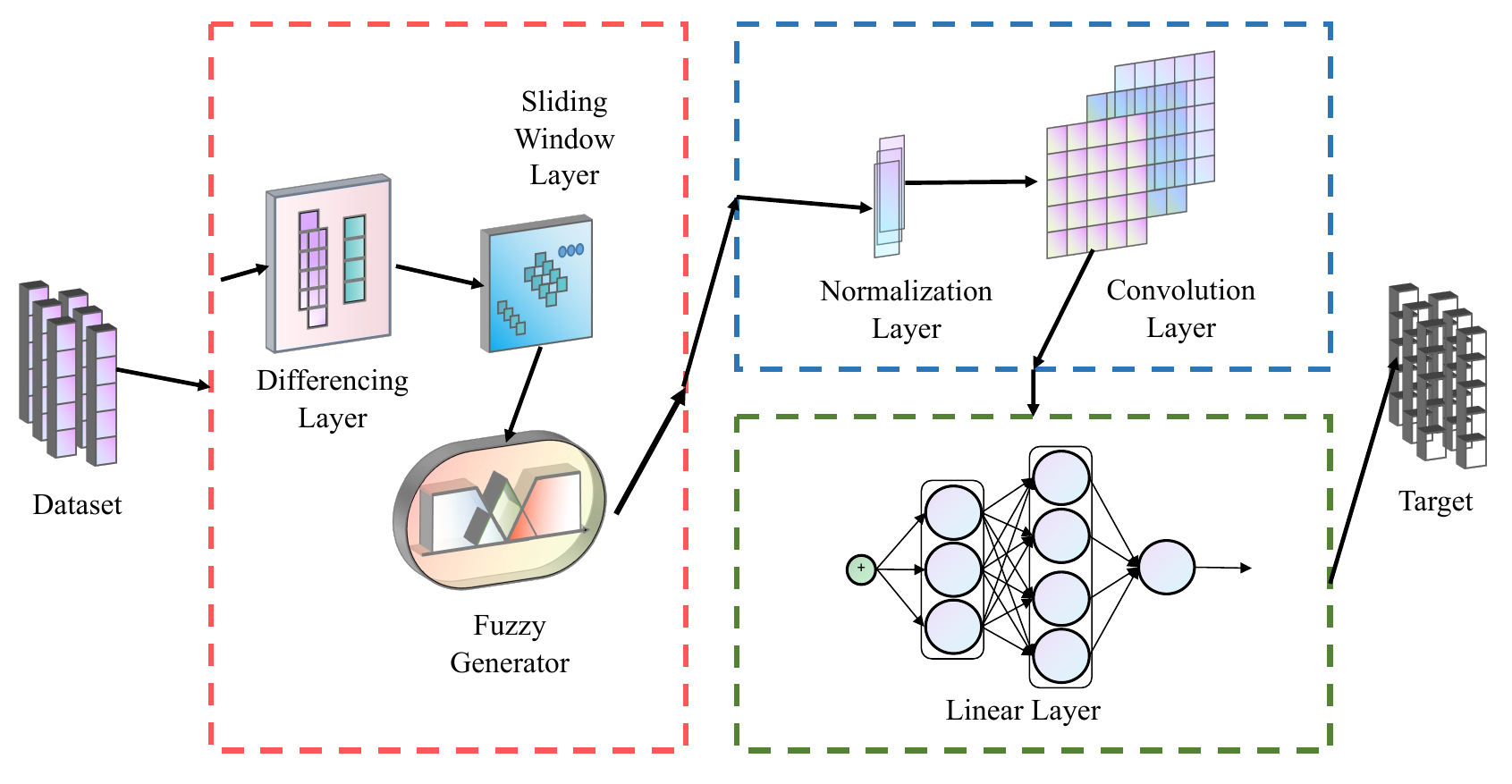}}
	\caption{Structure of Differential Fuzzy Convolutional Neural Network}
\end{figure*}

\section{Experiment and Analysis}
The datasets used in the experiment are M competition dataset and Construction Cost Index (CCI). Involved in the experiment compared methods from \textit{forecastingdata.org}. The experimental environment is AMD Ryzen 7950X (Cores: 16, Base Clock: 4.5GHz), RTX3090 (24GB) and the RAM is 64GB.

\subsection{Datasets}
The experiments involve 5 datasets with different sizes and types. 

\textit{Dataset 1: Construction Cost Index} 
\begin{itemize}
	\item \textit{Description:} The Construction Cost Index (CCI) is published once a month by Engineering Cost Record (ENR) \cite{ashuri2010time}. For the purpose of forecasting time series, 295 CCI data values (CCI data sets from January 1990 to July 2014) were employed. In the experiment, CCI divided the time series into training data sets and forecasting data sets in an 8:2 ratio.
	\item \textit{Intention:}  CCI data set is a single time series which is a \textbf{typical non-stationary time series} with significant trend. Fig.7 shows the trend of CCI test part. CCI data sets can verify the validity of DFCNN for non-stationary time series. In Fig.7, traditional FTS methods failed to forecast.
	\begin{figure}[htbp]%
		\centering
		\resizebox{\linewidth}{!}{
			\includegraphics{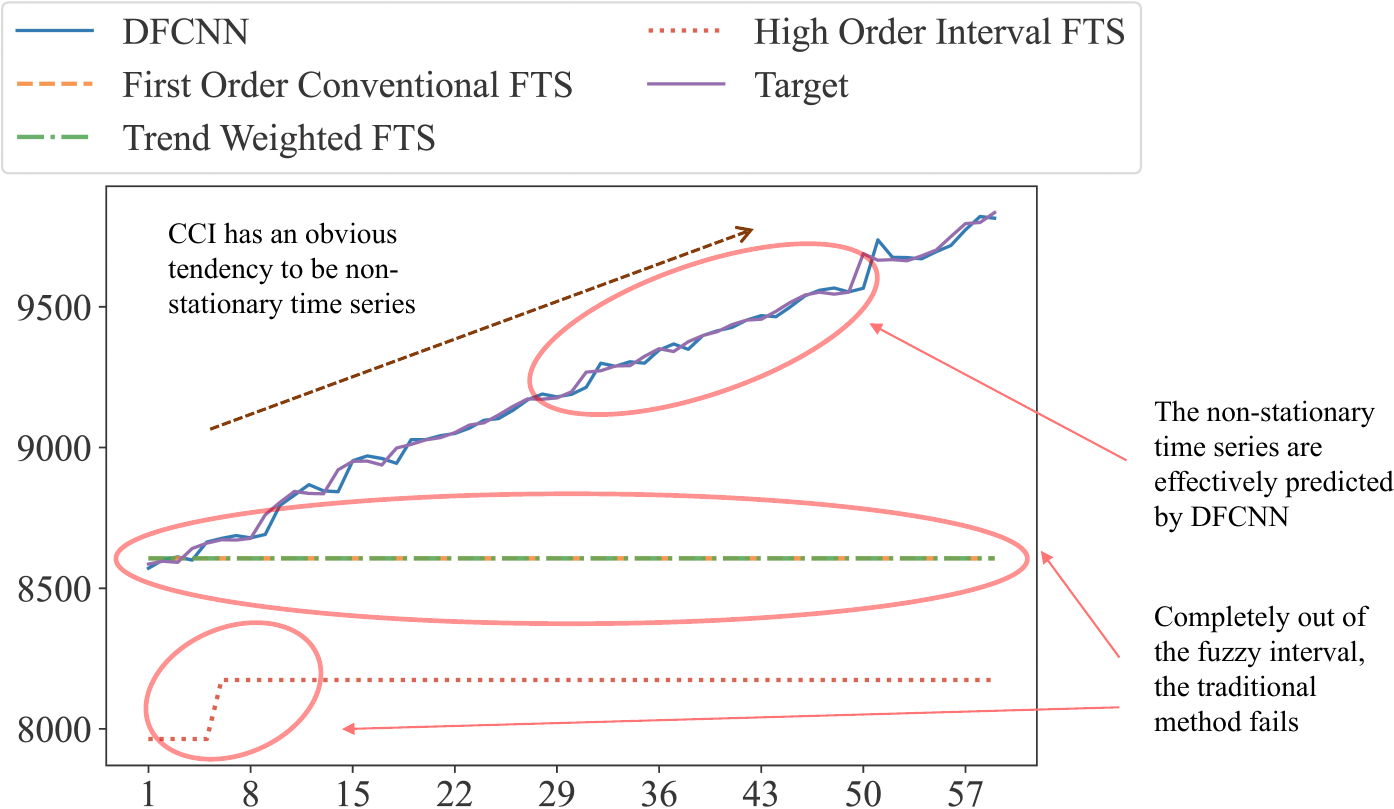}}
		\caption{Trend of CCI dataset and comparative results}
	\end{figure}
\end{itemize}

\textit{Dataset 2-5: Makridakis Competition}
\begin{itemize}
	\item \textit{Description of M Competition:} Makridakis Competition is also known as M Competition. The M Competition is an \textbf{open competition to evaluate the accuracy of different time series forecasting methods}. So M Competition can be used as a benchmark test of the effect of time series forecasting method.
	\item \textit{Dataset 2: M1 Competition} The first M Competition provided 1001 time series which is not real-time time series.
	\item \textit{Dataset 3: M2 Competition}  The second M Competition provided 29 time series which is real-time time series come from many collaborating organizations and competition announced in advance.
	\item \textit{Dataset 4: M3 Competition}  The third M Competition provided 3003 time series which were in the following domains: micro, industry, macro, finance, demographic, and other.The frequencies of time series include yearly, quarterly, monthly, daily, and \textbf{other}.
	\item \textit{Dataset 5: M4 Competition}  The fourth M Competition provided \textbf{100000 time series} which were in the following domains: micro, industry, macro, finance, demographic, and other. The frequencies of time series include yearly, quarterly, monthly, \textbf{weekly}, daily, and \textbf{hourly}.
	\begin{table}
		\centering
		\caption{M Dataset Features}
		\resizebox{0.9\linewidth}{!}{
			\begin{tblr}{
					cells = {c},
					cell{2}{1} = {r=4}{},
					cell{6}{1} = {r=4}{},
					cell{10}{1} = {r=5}{},
					cell{15}{1} = {r=7}{},
					hline{1-2,6,10,15,22} = {-}{},
					hline{5,9,14,21} = {2-6}{},
				}
				Dataset & Quantity & Number & Min & Max  & Horizion \\
				M1      & Yearly    & 181    & 15  & 58   & 6        \\
				& Quarterly & 203    & 18  & 114  & 8        \\
				& Monthly   & 617    & 48  & 150  & 18       \\
				& Total     & 1001   &     &      &          \\
				M2      & Quarterly  & 12     & 163 & 167  & 5     \\
				& Monthly1    & 42     & 33  & 94   & 14     \\
				& Monthly2      & 4      & 213 & 225  & 15    \\
				& Total     & 60     &     &      &          \\
				M3      & Yearly    & 645    & 20  & 47   & 6        \\
				& Quarterly & 756    & 24  & 72   & 8        \\
				& Monthly   & 1428   & 66  & 144  & 18       \\
				& Other     & 174    & 71  & 104  & 8        \\
				& Total     & 3003   &     &      &          \\
				M4      & Yearly    & 23000  & 19  & 841  & 6        \\
				& Quarterly & 24000  & 24  & 874  & 8        \\
				& Monthly   & 48000  & 60  & 2812 & 18       \\
				& Weekly    & 359    & 93  & 2610 & 13       \\
				& Daily     & 4227   & 107 & 9933 & 14       \\
				& Hourly    & 414    & 748 & 1008 & 48       \\
				& Total     & 100000 &     &      &          
		\end{tblr}}
	\end{table}
	\item \textit{Statistical characteristics:} Each M Competition specifies \textbf{different forecast horizons} for different frequency time series, in order to ensure that each prediction method has \textbf{the same prediction interval}. The maximum length of time series in M4 is \textbf{even longer than 8000 time points} which is a huge challenge for \textbf{single step forecasting methods} (include DFCNN and traditional FTS methods). Simultaneously, there is a large margin between minimum and maximum length of M datasets which \textbf{evaluates the capability of time series forecasting algorithms to extract predictive features from time series of varying lengths.}
	\item \textit{Intention:} M Competition datasets is involved by Monash Time Series Forecasting Repository (\textit {https://forecastingdata.org/}) which mentioned in the comparison model section and is \textbf{an open source} containing time series data set to assess benchmark \cite{godahewa2021monash}. M Competition datasets and Monash repository guarantee \textbf{the fairness of the data and the comprehensiveness of the evaluation of the time series forecasting algorithm.}
\end{itemize}

\subsection{Comparative Models}

In order to demonstrate the accuracy of the proposed model, the comparative models are divided into two categories: fuzzy time series forecasting method and Monash benchmark method.

\textit{Fuzzy time series forecasting method:}

Comparative FTSF methods demonstrate the improvement of DFCNN on previous work. In order to verify the advantages of DFCNN fairly, the comparative models is \textbf{representative backbone works of FTSF} because DFCNN is a novel FTSF method and also a backbone work.

\begin{itemize}
\item \textit{First Order Conventional FTS method (FOCFTS) \cite{chen1996forecasting}} is \textbf{a basic method} that transforms the crisp historical data into fuzzy sets based on predefined linguistic variables, establishes fuzzy logical relationships between the fuzzy sets, and converts the fuzzy forecasted values.

\item \textit{Trend Weighted FTS method (TWFTS) \cite{cheng2009forecasting}} extends the \textit{First Order Conventional FTS method} by incorporating trend analysis. It \textbf{detects the direction and magnitude of the changes} in the historical data and assigns trend weights to the fuzzy sets and rules.

\item \textit{High Order Interval FTS method (HOIFTS) \cite{severiano2017very}} extends the \textit{First Order Conventional FTS method} by incorporating weight algorithm and high order FTS algorithm. It utilized a \textbf{group of continuous fuzzy logic relationships} as input of FTSF algorithm and applied the weight algorithm to improve forecasting accuracy.

\item \textit{FTS with ANN (FTSWA) \cite{egrioglu_new_2009}} extends the \textit{First Order Conventional FTS method} by \textbf{determining fuzzy relationships by Artificial Neural Networks (ANN)}. After establishing the fuzzy sets, ANN processed the fuzzy relationships which solved the problem that observations of time series are defined linguistically or do not follow the assumptions required for time series theory.
\end{itemize}

A more intuitive difference between DFCNN and the comparative models is shown in Tab.2.

\begin{table*}[htbp]
	\centering
	\caption{The comparison between DFCNN and comparative models}
	\resizebox{\linewidth}{!}{
		\tabcolsep=1mm
	\begin{tabular}{lcccccc}	
		\hline
		Model & Weighted & Learnable & Non-stationary & Expanded-Relationship & Differences with DFCNN  \\ \hline
		FOCFTS & $\times$ & $\times$ & $\times$ &$\times$ & Same as below  \\ 
		TWFTS & \checkmark & $\times$ & $\times$ & $\times$ & DFCNN uses differences algorithm to remove trends  \\ 
		HOIFTS & \checkmark & $\times$ & $\times$ & \checkmark & DFCNN extends the length of fuzzy relationships  \\ 
		FTSWA & $\times$ & \checkmark & $\times$ & $\times$ & DFCNN learns before establishing fuzzy sets  \\ 
		\textbf{DFCNN} & \checkmark & \checkmark & \checkmark &\checkmark & -  \\ \hline
	\end{tabular}}
\end{table*}

The Monash Time Series Forecasting Benchmark is aimed to \textbf{compare common models of DFCNN and non-FTS with M Competiton datasets}. These include statistical models such as \textit{Auto-Regressive Integrated Moving Average}, machine learning models such as \textit{CatBoost}, deep learning models such as \textit{Transformer} and so on.

\textit{Monash Time Series Forecasting Benchmark:}
\begin{itemize}
	\item \textit{Exponential Smoothing (ETS) \cite{hyndman2008forecasting}}  is a time series forecasting method that uses an exponential window function to smooth data. Unlike the simple moving average, where past observations are weighted equally, exponential functions are used to assign exponentially decreasing weights over time. 
	
	\item \textit{Dynamic Harmonic Regression Auto-Regressive Integrated Moving Average (ARIMA) \cite{hyndman2018forecasting}} is a forecasting model that combines the strengths of both global and local models. It leverages the strengths of Dynamic Harmonic Regression and ARIMA (Auto-Regressive Integrated Moving Average) models to provide accurate forecasts 
	
	\item \textit{Pooled Regression Model (PR) \cite{trapero2015identification}} is a type of model that has constant coefficients, referring to both intercepts and slopes. It is usually carried out when we have available time series of cross-sections such as data that has observations over time for several different groups or cross-sections.
	
	\item \textit{CatBoost \cite{prokhorenkova2018catboost}} is not a specific time series forecasting model. However, it can be used to forecast time series data by using its regression or classification models. CatBoost is a high-performance open-source library for gradient boosting on decision trees. It is developed by Yandex researchers and engineers.
	
	\item \textit{DeepAR \cite{salinas2020deepar}} is a probabilistic forecasting model that uses an autoregressive Recurrent Neural Network (RNN) to forecast scalar (one-dimensional) time series. It combines deep learning with traditional probabilistic forecasting by using a RNN  in conjunction with an Auto-Regressive (AR) process to make predictions.
	
	\item \textit{WaveNet \cite{borovykh2017conditional}} is a deep generative model of raw audio waveforms developed by DeepMind. While it has been primarily used for generating speech and audio, it has also been applied to time series forecasting. 
	
	\item \textit{Transformer \cite{vaswani2017attention}} is originally developed for natural language processing tasks, such as machine translation, but has since been applied to a wide range of tasks, including time series forecasting. The Transformer is based on the self-attention mechanism, which allows it to weigh the importance of different parts of the input sequence when making predictions.
\end{itemize}

\subsection{Basic Experiment Setting}

Mean Absolute Error (MAE) \cite{godahewa2021monash} are selected as error indexes and the definition shown in Eq.33.

\begin{equation}
	MAE=\frac{1}{N}\sum_{t=1}^{N}\left|\hat{y}(t)-y(t)\right|
\end{equation}

\textit{Error Calculation Setting:}
\begin{itemize}
	\item \textit{CCI Dataset:} On the test set of \textit{CCI}, all models are predicted in a single step continuously, and finally the error is calculated.
	
	\item \textit{M Competition Dataset:} \textit{M Competion Datasets} contain not only one time series in each sub-dataset. So all models are predicted in a single step continuously, and \textbf{save the mean MAE error} of all tested time series in each sub-dataset.
	
\end{itemize}

As mentioned in Section IV, \textit{DFCNN} keeps $lookback=2$ which is default parameter for $lookback$ because the length of fuzzy relationship established by comparative \textit{FTSF} models is 2 and \textit{DFCNN} set the same length of fuzzy relationship established in learning process as Eq.30. Tab.3 shows the detailed parameters of \textit{DFCNN} used in experiment 1 and experiment 2.

\begin{table}[htbp]
	\centering
	\caption{Experiment Parameter Details}
	\resizebox{\linewidth}{!}{
		\begin{tabular}{ccc}
			\hline
			\textbf{Parameter} & \textbf{Value}    & \textbf{Note} \\ \hline
			seed               & $3407$                 & /  \\
			lookback           & $2$                 & Default value \\
			out                & $2$                 & Default value \\
			Training Epoch     & $100$               &        /       \\
			Loss Function      & MAE          &    /           \\
			Optimizer          & NAdam             &      /         \\
			Scheduler          & ReduceLROnPlateau &   /            \\
			Scheduler Target   & Current Loss      &       /        \\
			Learning Rate   & $10^{-2}$      &       /        \\ \hline
		\end{tabular}
	}
\end{table}

The implementation framework for \textit{DFCNN} is PyTorch. For comparative FTSF models, they all implemented by pyFTS Library and set $npart=15$ which is the maximum values on CCI training dataset which keeps that \textbf{fuzzy interval is divided into most fuzzy intervals}. For Monash Time Series Forecasting Benchmark, \textbf{the implementation is provided by Monash Time Series Forecasting Repository (\textit {https://forecastingdata.org/})} which has already been \textbf{set uniform parameter for each dataset and comparative models} and comparative models is implemented by GluonTS and R Language.

\subsection{Experiment 1: Comparison Between The Proposed Method And Previous Fuzzy Time Series Forecasting Methods}

\begin{figure*}[htbp]%
	\centering
	\resizebox{0.8\linewidth}{!}{\includegraphics{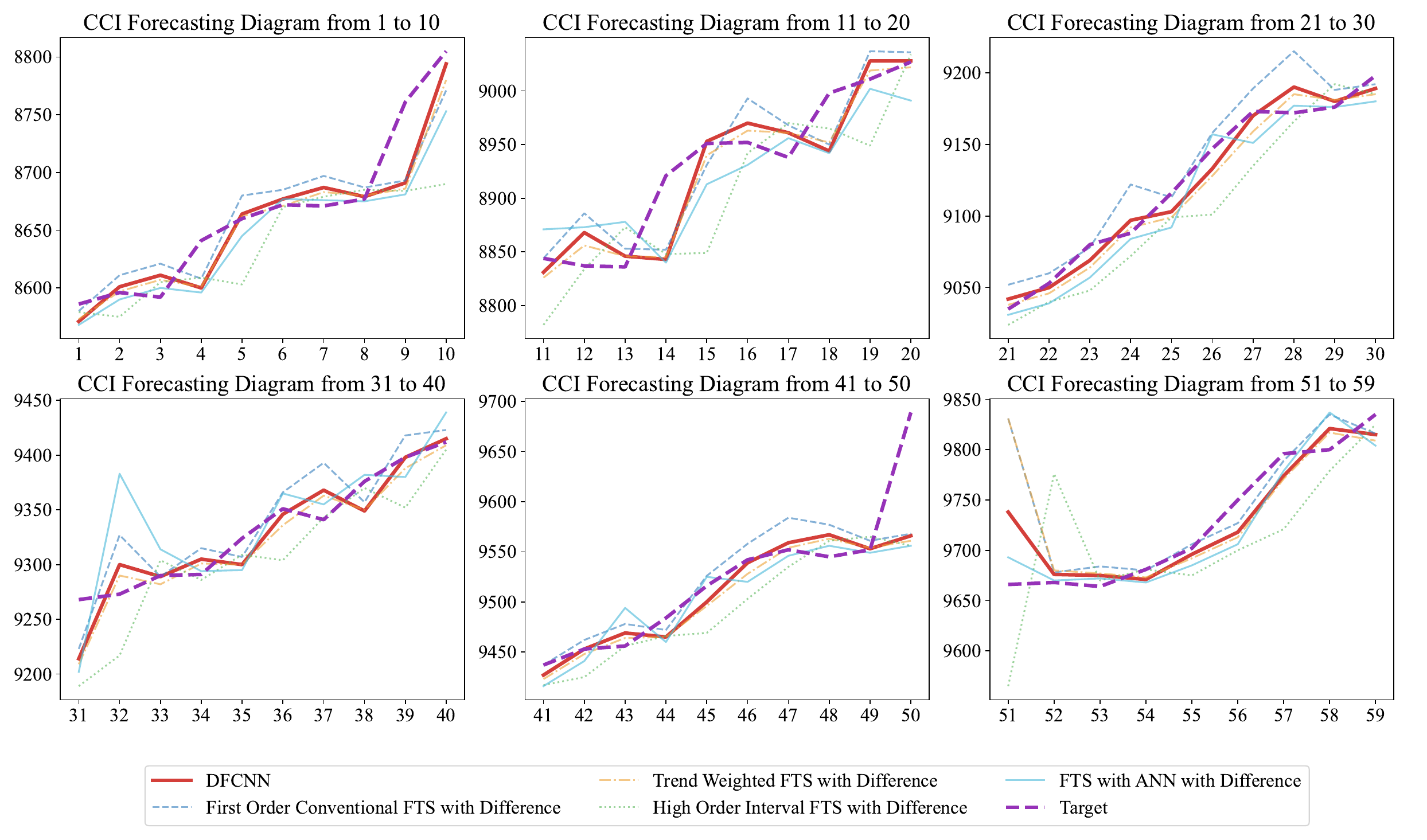}}
	\caption{Forecasting line  diagram of comparative experiment 1 }
\end{figure*}

 Many academics in the construction sector have undertaken study using the CCI data. There is a significant issue that CCI is type of economic index which means it will increase with economic development. Therefore, the non-stationary characteristic of CCI need to be considered into forecasting modeling.\textbf{ Most \textit{FTSF} ignores the non-stationary characteristic} because \textbf{when fuzzy sets of model are established, the boundaries of fuzzy sets are specific} which will not change with the trend of time series and lead to forecasting failure in Fig.7. 
 
 \begin{table}[htbp]
 	\centering
 	\caption{forecasting MAE error on the CCI dataset}
 	\resizebox{\linewidth}{!}{
 		\tabcolsep=5mm
 		\begin{tabular}{ccc}
 			\hline
 			Method (with Difference)      & MAE             \\ \hline
 			High Order Interval FTS       & 31.92 $(\uparrow 40.32\%)$  \\
 			FTS with ANN        &27.84 $(\uparrow 31.57 \%)$                                   \\
 			First Order Conventional FTS  & 25.74 $(\uparrow 26.11\%)$               \\
 			Trend Weighted FTS            & 21.39 $(\uparrow11.08\%)$                   \\
 			\textbf{DFCNN (default)}      & \textbf{19.05}  \\ \hline
 	\end{tabular}}
 \end{table}
 
 To ensure the feasibility of comparative models, the same difference algorithm is applied on the comparative models. Tab.4 shows the forecasting error of CCI dataset. Fig. 8 is forecasting line diagram of experiment 1.
 
 In experiment, \textbf{the difference between \textit{DFCNN} and comparative is process of fuzzy relationships. } In Tab.2, the differences between model optimization methods are compared. 
 
 \begin{itemize}
 	\item  \textit{First Order Conventional FTS} is regarded as basic model. Optimal rate $\eta$ is a feature to describe how much better proposed model predict than the comparison model in Eq.34. $e_c$ and $e_p$ stand for forecasting error of comparative model and proposed model. Compared to \textit{DFCNN},  \textit{First Order Conventional FTS} The boost of \textit{DFCNN} relative to \textit{First Order Conventional FTS} is huge which optimal rate reaches up to \textbf{26.11\%}. 
 	
 	\begin{equation}
 		\eta = \frac{e_c-e_p}{e_c}\times100\%
 	\end{equation}
 	
 	\textit{Trend Weighted FTS}, \textit{High Order Interval FTS}, \textit{High Order Interval FTS} and \textit{DFCNN} are all opitimal methods based on \textit{First Order Conventional FTS}. \textbf{But \textit{First Order Conventional FTS} is not farthest one from the target line in Fig.8.} The following focuses on the improvement ways of \textit{First Order Conventional FTS} and differences between \textit{DFCNN} and comparative model for FTSF.
 	
 	\item  \textit{Trend Weighted FTS} considered the trend of FTS, and it is \textbf{feasible when forecasted time series are stationary}. So for CCI dataset,  \textit{Trend Weighted FTS} handles the all parts of CCI with frequent trend changes well, so optimal rate is the lowest one \textbf{19.05\%}. \textit{DFCNN} contains difference algorithm which is also a approach to \textbf{ensure the time series made difference are more possible to be close to be stationary}. Therefore, the performance of \textit{DFCNN} in processing trend changes is similar to that of \textit{Trend Weighted FTS}, which is close to the target line. Another improve way of FTSF in \textit{Trend Weighted FTS} is weighted operation which is a process in expert system. \textit{DFCNN} utilizes Linear Layer to weight Fuzzy Token which equals to fuzzy relationships in \textit{Trend Weighted FTS}.

 	\item   \textit{High Order Interval FTS} considered a group of continuous fuzzy logic relationships as input which expand the original fuzzy logic in First Order Conventional FTS and contains more information. \textbf{More accurate information helps First Order Conventional FTS to fetch lower error, but more meaningless information leads to higher error}. \textit{DFCNN} also considers more information and controls the information by parameter $lookback$. Similar to First Order Conventional FTS, \textbf{\textit{DFCNN} expands the length of fuzzy relationships} instead of expanding the input of fuzzy relationships which both increase the quantity of fuzzy information. In CCI dataset, \textit{High Order Interval FTS} has the \textbf{worst performance} $MAE=31.92$ and highest optimal rate \textbf{40.32\%} which means that the improvement with \textit{DFCNN} is greatest on the basis of \textit{High Order Interval FTS}, especially in terms of expanding the relationship. In Fig.8 time 51 and 52, \textit{High Order Interval FTS} even loses the fiducial standard of last value. The problem is that \textbf{specific weight or established fuzzy relationships is much possible to be different with time series' actual variation}. Unlike \textit{First Order Conventional FTS} and \textit{Trend Weighted FTS}, the input to \textit{High Order Interval FTS} is a set of fuzzy relationships. When \textit{High Order Interval FTS} cannot establish a group of valid weight for the input data, \textit{High Order Interval FTS} will accumulate errors of each fuzzy relationship and its weight. For a initial \textit{DFCNN}, it can not solve accumulating errors, but learnable feature can do. With weight of model updating, weight of \textit{DFCNN} tends to fit to input time series CCI. 
 	
 	\item \textit{FTS with ANN} is hybrid model which only utilizes ANN to reason fuzzy relationship. Comparing to \textit{First Order Conventional FTS}, the only difference is the way of fuzzy reasoning. \textit{DFCNN} is not similar to \textit{FTS with ANN}. Fuzzy Token records the values of time series and fuzzy interval. But the actual fuzzy relationship is not established, equivalent information could be learning by Convolution Layer. \textit{FTS with ANN} does not perform well in CCI ($MAE=27.84$ and $\eta=31.57\%$ ) \textbf{due to time 32 in Fig.8}. \textit{FTS with ANN} outputs single fuzzy relationship and defuzzifies the fuzzy relationship which means \textbf{there is more possible to generate a bias of fuzzy relationship than weighted models} such as \textit{Trend Weighted FTS and DFCNN}. \textbf{The robustness of \textit{FTS with ANN} is weak}. \textit{FTS with ANN} is relatively accurate in predicting other parts of CCI, indicating that learning ability is of great help to enhance the prediction accuracy of time series.
 	
 \end{itemize}
 
 Many financial data, such as stocks and indices, are similar to CCI in that they are non-stationary and changeable. \textit{DFCNN} combines the advantages of comparison model and shows good results in trend and model construction. At the same time, compared with large predictive models, \textit{DFCNN} has a low time complexity, which will be analyzed in detail in the time complexity analysis. However, \textit{DFCNN} is a single step prediction model. \textit{DFCNN} is not suitable for making continuous predictions with short intervals. \textbf{For example, DFCNN is suitable for predicting the highest price of a stock the next day, but is not suitable for predicting the trend of a stock in real time.} \textbf{The future derived model of \textit{DFCNN} can focus on how to realize multi-step forecasting}, which needs to give up the difference algorithm and use other methods to remove the trend of time series.

\subsection{Time Complexity Analysis}

\textit{DFCNN} is a lightweight and sophisticated forecasting model. Assume there are $n$ elements in a time series. Difference Layer and Sliding Windows Layer 's time complexity is $O(n)$. The time complexity of the computation in the Fuzzy Generator Layer is $O(1)$ for each time point, and there is no recursion or looping. So the time complexity of Fuzzy Generator Layer is $O(n)$. Batch Norm Layer, Convolution Layer (one dimension) and Linear Layer are common modules in the neural network and their time complexity is $O(n)$. What needs attention is that the input channels of Convolution Layer affects the time complexity. Usually, it is assume that $lookback << n$. If $lookback << n$ is not established, Convolution Layer's time complexity is $O(n^2)$. In summary, time complexity of DFCNN usually approaches to $O(n)$ and tends to $O(n^2)$ when $lookback << n$ is not established.

\begin{table*}
	\centering
	\caption{Forecasting \textbf{mean} MAE error of DFCNN with Default Settings}
	\resizebox{0.9\linewidth}{!}{
	\begin{tblr}{
			cells = {c},
			cell{2}{9} = {fg=red},
			cell{3}{9} = {fg=red},
			cell{4}{9} = {fg=red},
			cell{5}{7} = {fg=red},
			cell{5}{9} = {fg=green},
			cell{6}{9} = {fg=red},
			cell{7}{9} = {fg=red},
			cell{8}{9} = {fg=red},
			cell{9}{9} = {fg=red},
			cell{10}{9} = {fg=red},
			cell{11}{9} = {fg=red},
			cell{12}{9} = {fg=red},
			cell{13}{9} = {fg=red},
			cell{14}{9} = {fg=red},
			cell{15}{9} = {fg=red},
			cell{16}{9} = {fg=red},
			cell{17}{9} = {fg=red},
			hline{1-2,18} = {-}{},
			stretch = 0.8
		}
		& ETS       & ARIMA     & PR        & CatBoost  & DeepAR    & WaveNet           & Transformer & \textbf{DFCNN}     \\
		M1 Yearly    & 146110.11 & 145608.87 & 134246.38 & 215904.20 & 152084.40 & 284953.90         & 164637.90   & \textbf{66599.77}  \\
		M1 Quarterly & 2088.15   & 2191.10   & 1630.38   & 1802.18   & 1951.14   & 1855.89           & 1864.08     & \textbf{1197.90}   \\
		M1 Monthly   & 1905.28   & 2080.13   & 2088.25   & 2052.32   & 1860.81   & 2184.42           & 2723.88     & \textbf{1559.51}   \\
		M2 Monthly1  & 42539.44  & 46204.07  & 34909.65  & 64139.56  & 33925.16  & \textbf{31885.42} & 32913.58    & \textbf{125772.73} \\
		M2 Monthly2  & 115955.49 & 136663.12 & 105961.92 & 99467.05  & 126581.94 & 112061.01         & 118304.19   & \textbf{34484.96}  \\
		M2 Quarterly & 16.84     & 13.73     & 14.42     & 48.55     & 21.92     & 20.55             & 16.22       & \textbf{8.96}      \\
		M3 Yearly    & 1031.40   & 1416.31   & 1018.48   & 1163.36   & 994.72    & 987.28            & 924.47      & \textbf{525.81}    \\
		M3 Quarterly & 513.06    & 559.40    & 519.30    & 593.29    & 519.35    & 523.04            & 719.62      & \textbf{334.94}    \\
		M3 Monthly   & 626.46    & 654.80    & 692.97    & 732.00    & 728.81    & 699.30            & 798.38      & \textbf{520.00}    \\
		M3 Other     & 194.98    & 193.02    & 234.43    & 318.13    & 247.56    & 245.29            & 239.24      & \textbf{92.24}     \\
		M4 Yearly    & 920.66    & 1067.16   & 875.76    & 929.06    & -         & -                 & -           & \textbf{413.74}    \\
		M4 Quarterly & 573.19    & 604.51    & 610.51    & 609.55    & 597.16    & 596.78            & 637.60      & \textbf{382.43}    \\
		M4 Monthly   & 582.60    & 575.36    & 596.19    & 611.69    & 615.22    & 655.51            & 780.47      & \textbf{343.66}    \\
		M4 Weekly    & 335.66    & 321.61    & 293.21    & 364.65    & 351.78    & 359.46            & 378.89      & \textbf{222.52}    \\
		M4 Daily     & 193.26    & 179.67    & 181.92    & 231.36    & 299.79    & 189.47            & 201.08      & \textbf{62.64}     \\
		M4 Hourly    & 3358.10   & 1310.85   & 257.39    & 285.35    & 886.02    & 393.63            & 320.54      & \textbf{141.87}    
	\end{tblr}}
\end{table*}

\subsection{Experiment 2: Comparison Between The Proposed Method And Common Time Series Forecasting Methods}

\begin{figure}[htbp]%
	\centering
	\resizebox{0.7\linewidth}{!}{
		\includegraphics{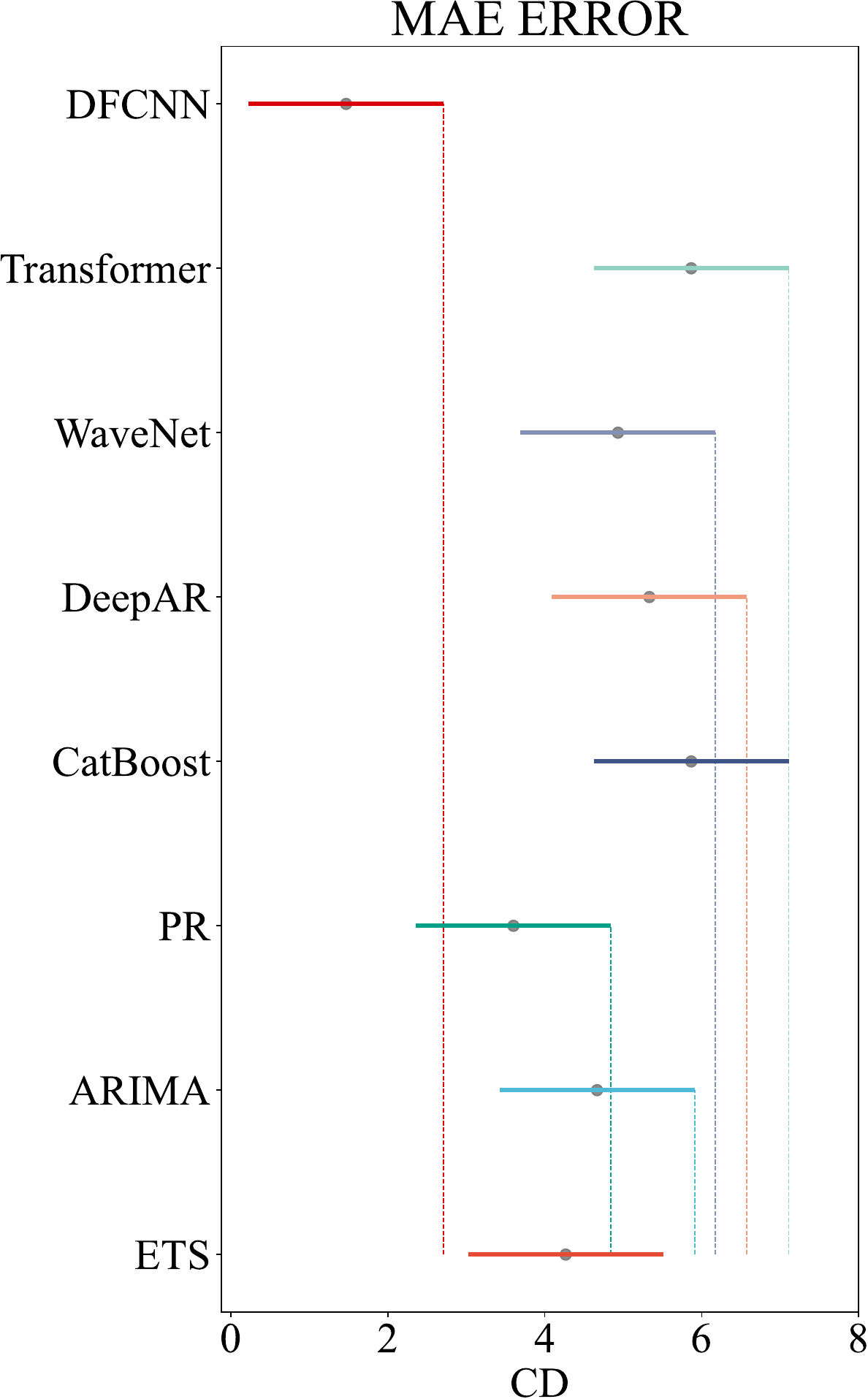}}
	\caption{Friedman test figure of DFCNN and compared methods in M competition dataset ($\alpha=0.1$) }
\end{figure}

\begin{figure*}[htbp]%
	\centering
	\subfloat[]{
		\includegraphics[width=0.47\linewidth]{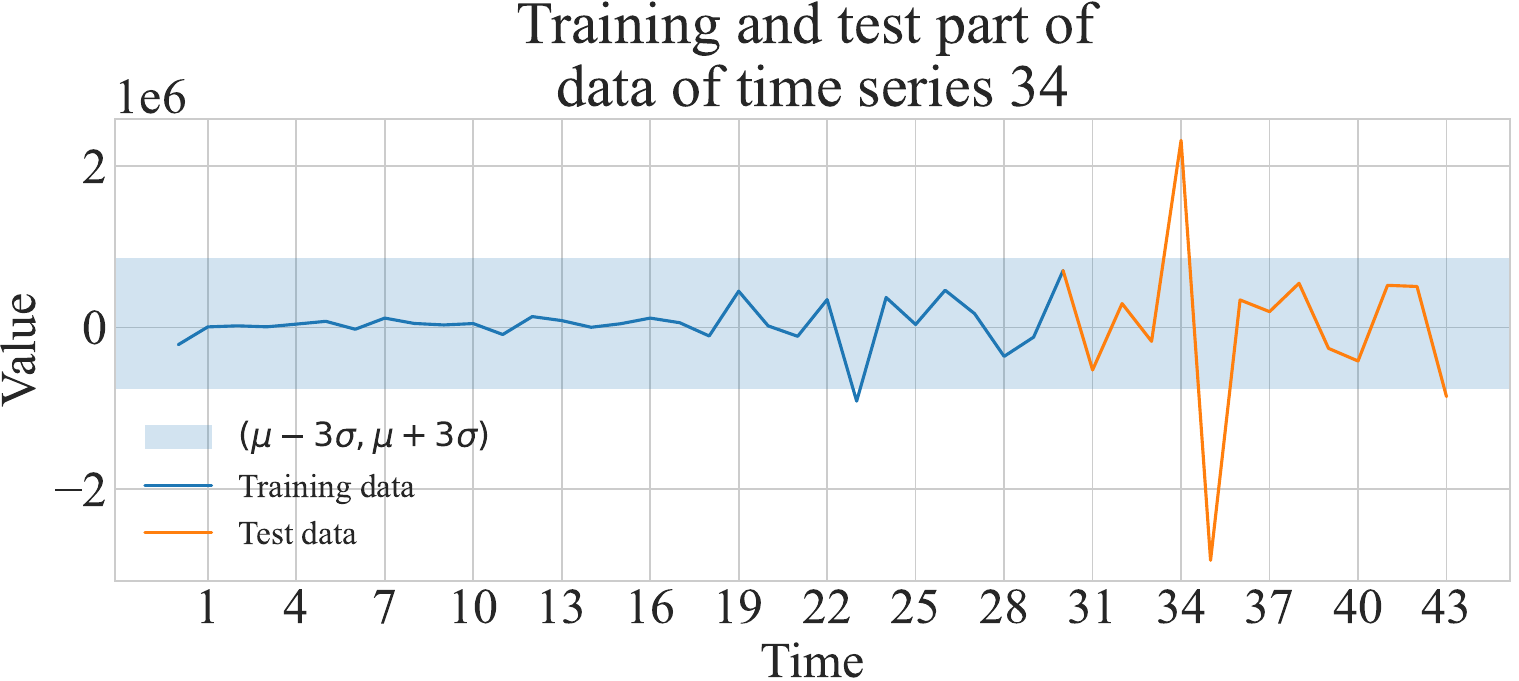}
	}
	\subfloat[]{
		\includegraphics[width=0.47\linewidth]{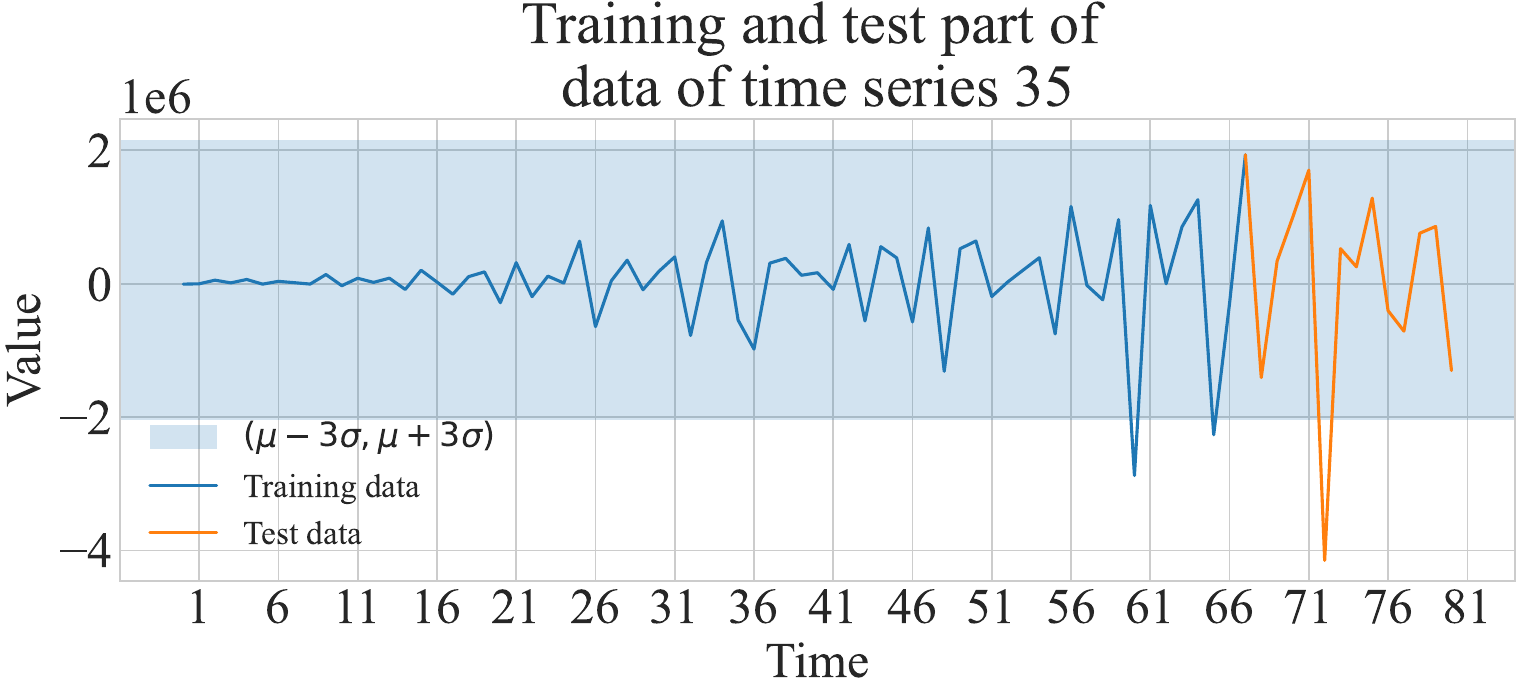}
	}\\ 
	\caption{Abnormal time series 34, 35 in M2 Monthly1}
\end{figure*}

In Tab.5, there is \textbf{mean MAE} error of DFCNN and comparison model and the red marked is the lowest error in the single dataset. In order to critically assess the difference in prediction effectiveness between DFCNN and other methods, these methods were analyzed quantitatively by applying the Nemenyi test ($\alpha = 0.1$). Fig. 9 is Friedman test figure. The horizontal line segment of DFCNN has no overlapping area with most methods, and it can be concluded that DFCNN significantly outperforms the vast majority of compared algorithms.  Although \textit{PR} horizontal line segment has overlapping area with DFCNN, the length of overlapping area is \textbf{so short that could be ignored}.

\begin{figure}[htbp]%
	\centering
	\resizebox{0.8\linewidth}{!}{
		\includegraphics{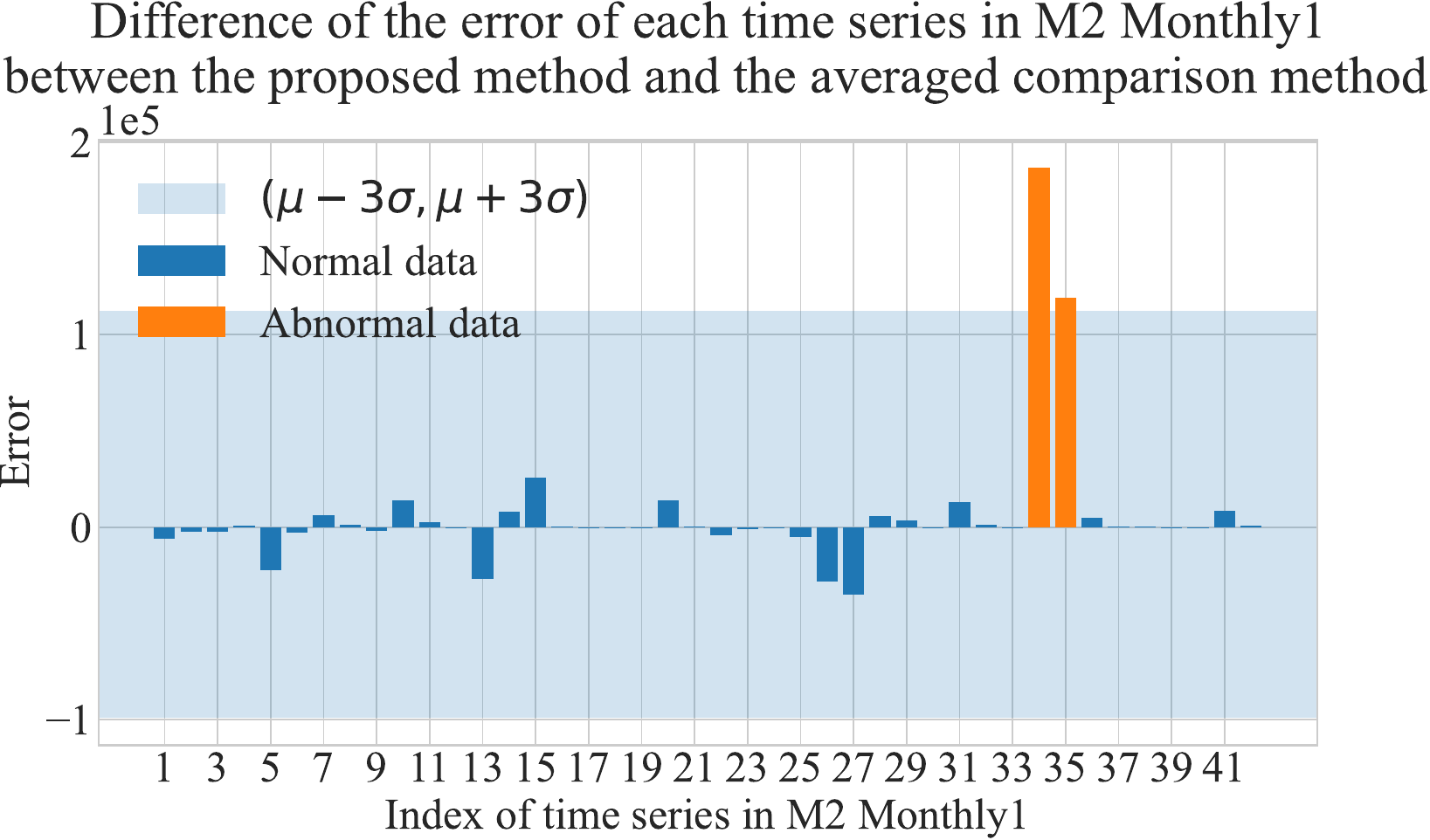}}
	\caption{DFCNN forecasts abnormally in Time series 34, 35 of M2 Monthly1 }
\end{figure}

 DFCNN gains all red marks except red mark in M2 Monthly1. DFCNN dose not handle the forecasting well and gains the unsatisfactory error which marked green. The phenomenon is abnormal, so the difference between DFCNN and averaged comparison method of each time series in M2 Monthly1 is depicted as bar chart in Fig.11. To figure out the abnormal errors, the errors are applied \textbf{Z Test}. The average of errors is recorded as $\mu$ and the standard difference of errors is recorded as $\sigma$. According to Gaussian distribution, there is 99.7\% probability that the errors out of $(\mu-3\sigma, \mu+3\sigma)$ (blue area in Fig.11)  are abnormal. Obviously, errors of time series 34,35 marked orange are abnormal. In the forecasting part of time series 34,35, there is a large margin raising and descent which is different from training part. The blue area in Fig.10 (a) and Fig.10 (b) is Z Test area which is 99.7\% probability that the test data out of $(\mu-3\sigma, \mu+3\sigma)$  are abnormal. To be different, the Z Test only considers the average value $\mu$ and standard difference $\sigma$ of training data. \textbf{\textit{DFCNN} cannot learn these abnormal data efficiently}, which leads to forecasting errors on M2 Monthly2.
 
 The following compares \textit{DFCNN} with other models for model categories:
 \begin{itemize}
 	\item \textit{Statistical Models:} \textit{ETS}, \textit{ARIMA} and \textit{PR} belong to statistical model. Through Nemenyi test in Fig.9, it can be concluded that the statistical model has a good effect on the M Competition datasets. The forecasting goal of M Competition datasets in experiment 2 is short sequence single step forecasting, because the forecasting step sizes defined by M Competition datasets are all short. The forecasting of statistical models come entirely from the statistical features of the data, so they can be relatively accurate even with small amounts of data. \textit{DFCNN} can also handle small amounts of data well. Fuzzy modeling and Machine learning is different from deep learning in that fuzzy modeling is based on reasoning of previously established fuzzy relationships. \textit{DFCNN} can also be understood as \textbf{an FTS model with more flexible parameters or weights}. \textit{DFCNN} can learn to make the parameters more suitable for the target data, so the forecasting is smaller than that of the statistical model.
 	\item \textit{Machine Learning Model:} \textit{Catboost} belongs to machine learning model. The essential difference between \textit{DFCNN} and \textit{Catboost} is that \textit{Catboost} is not based on a specific forecasting process and \textit{DFCNN} is based on \textit{FTSF}. Therefore, Catboost performs poorly on small data sets such as M1, M2, M3 and other models. DFCNN maintains the ground forecasting error in different data sets. \textbf{When the amount of data increases and \textit{FTSF} is not suitable for processing, \textit{DFCNN's} learning parameters play a role in this case.}
 	\item \textit{Deep Learning Models:} \textit{DeepAR}, \textit{WaveNet} and \textit{Transformer} belong to deep learning model. \textit{DeepAR} is a hybrid model that combines \textit{AR} and \textit{RNN} models. \textit{AR} can predict time series with small amount of data and modify the result through \textit{RNN} network. However, when the amount of data increases, for example, in M4 Hourly data sets, \textit{AR} as a statistical model cannot effectively model data, resulting in a sharp drop in the accuracy rate of \textit{DeepAR}. WaveNet itself can be used for data generation, so \textit{WaveNet} captures data differently than regression models. \textit{WaveNet} learns more abstract features, and the benefit is that it handles abnormal data better than regression models. This feature was demonstrated in the experimental results of M2 Monthly1. \textit{Transformer} introduces the Attention mechanism and is a common backbone.\textit{ Transformer} focuses on interrelationships before processing underlying information, such as modeling natural language. Therefore, \textbf{\textit{Transformer} will have large forecasting error due to insufficient data on small data. This problem is also common to most deep time series forecasting models.} For multi-step forecasting tasks, \textit{Transformer} will perform better, including \textit{RNN}. In the task of single-step forecasting, \textit{DFCNN} has achieved good performance in various data quantities.
 \end{itemize}
 
 In short, compared with statistical model, \textit{DFCNN} can guarantee the stability of forecasting with small data volume. Compared with machine learning and deep learning models, \textit{DFCNN}'s learning ability to optimize weights can gain advantages in single-step forecasting of large data volumes, which is not possible with statistical models and traditional \textit{FTSF}. At the same time, the underlying \textit{FSTF} construction in \textit{DFCNN} will not cause the model forecasting to be inaccurate due to insufficient data. Therefore, \textbf{\textit{DFCNN} is suitable for single-step time series forecasting of various lengths with little abnormal data.}

\subsection{Experiment 3: Hyperparameter Testing Of The Model}

\begin{figure*}[htbp]%
	\centering
	\subfloat[]{
		\includegraphics[width=0.23\linewidth]{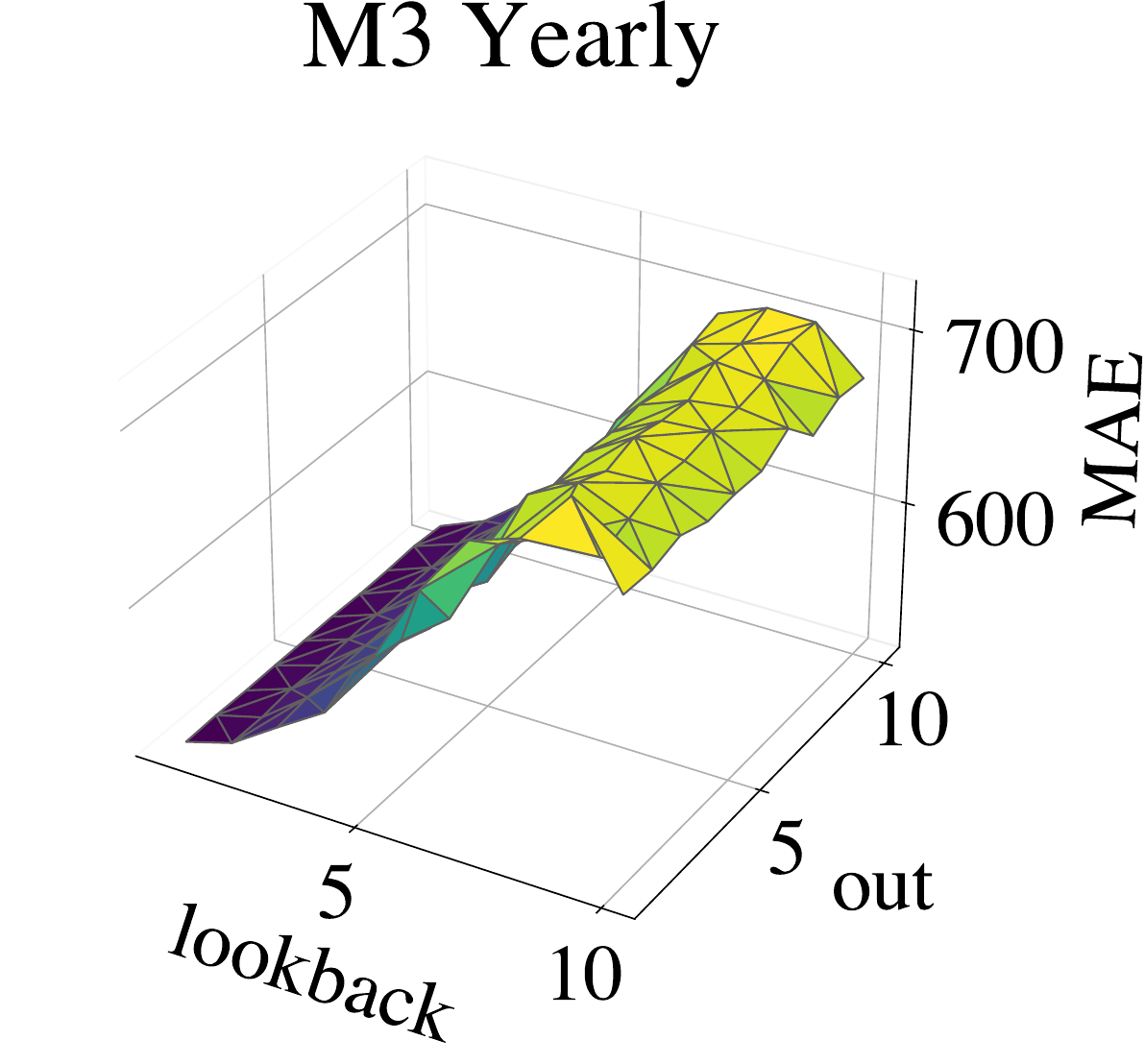}
	}\hfill
	\subfloat[]{
		\includegraphics[width=0.23\linewidth]{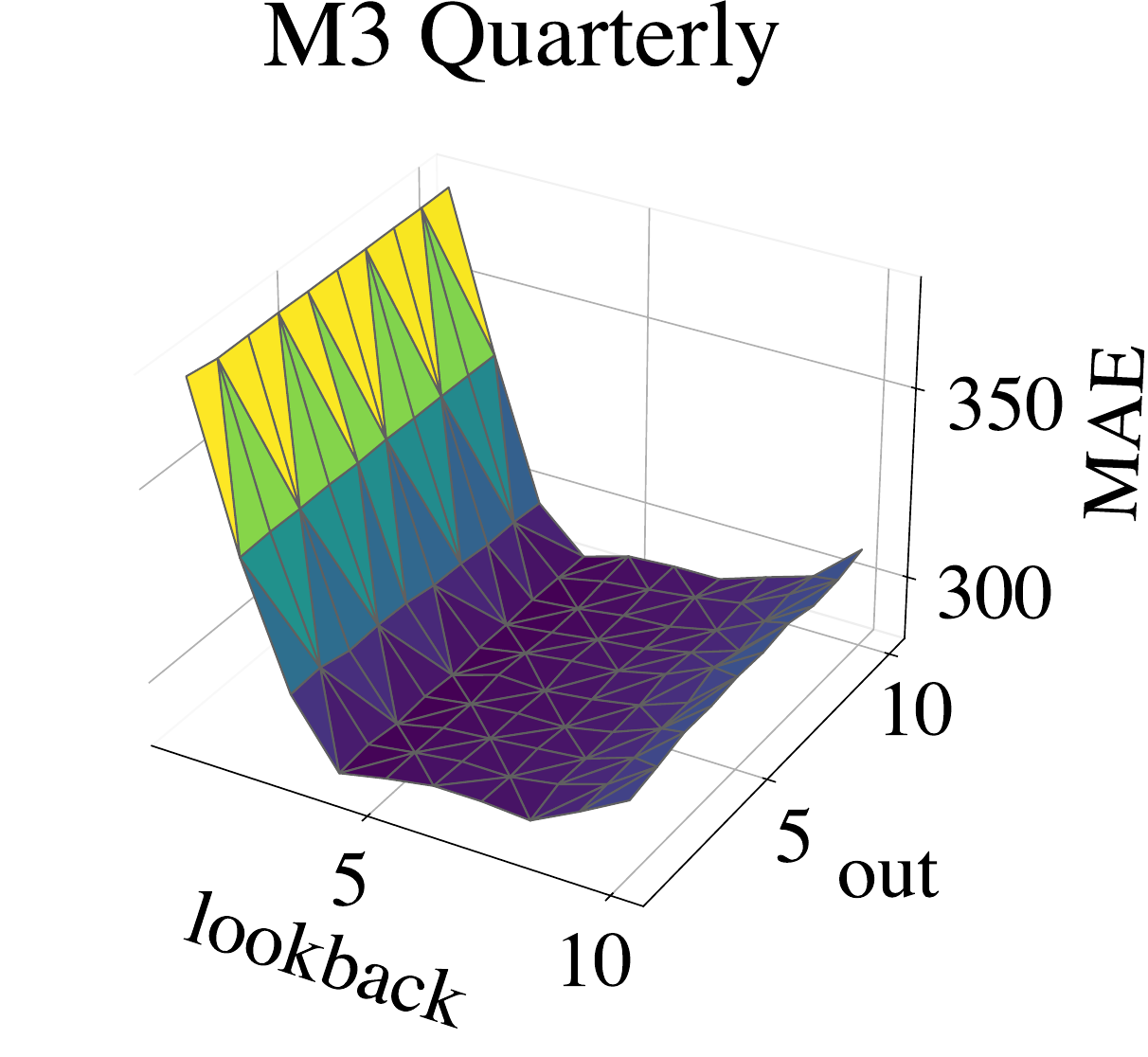}
	}\hfill
	\subfloat[]{
		\includegraphics[width=0.23\linewidth]{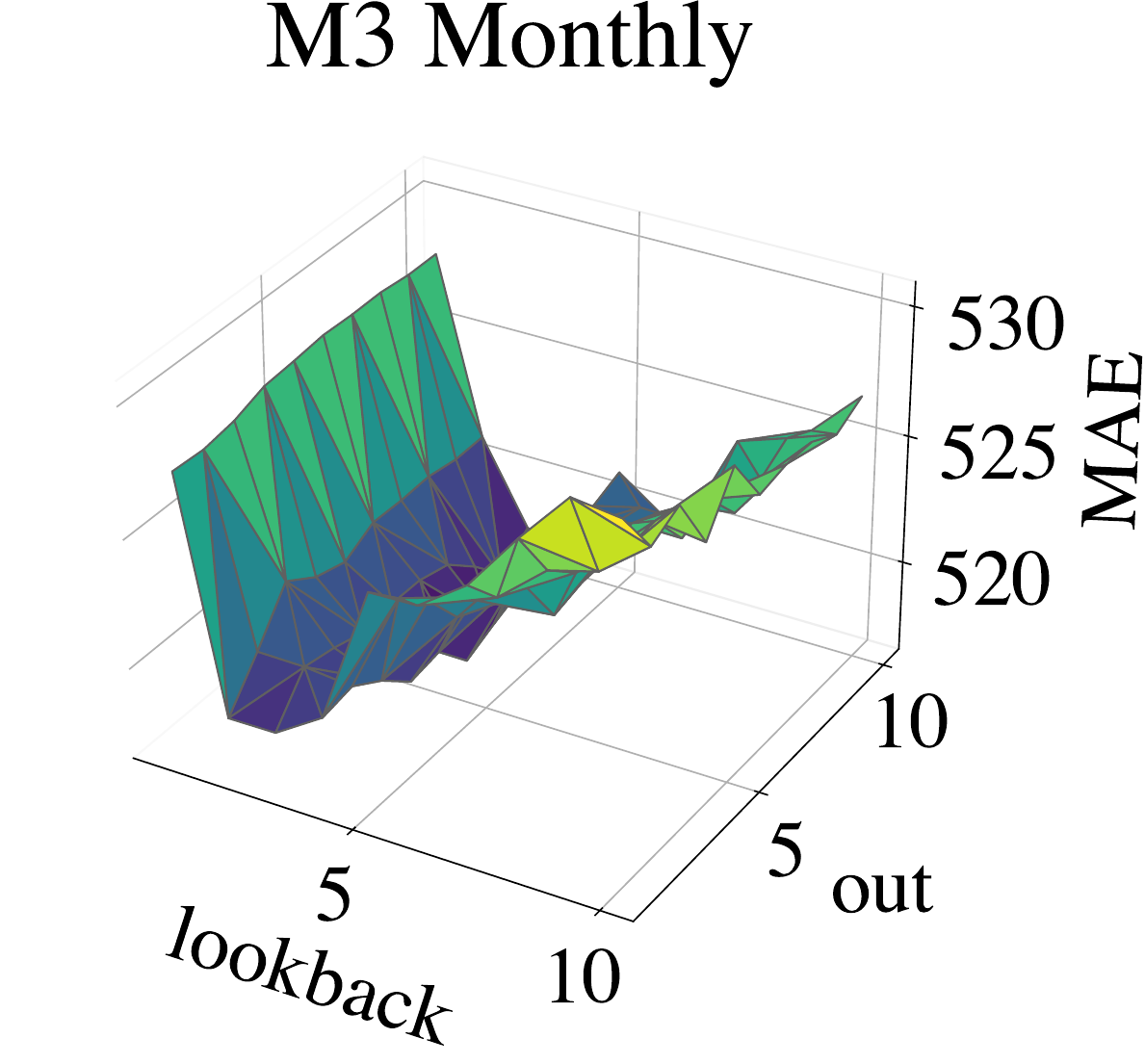}
	}\hfill
	\subfloat[]{
		\includegraphics[width=0.23\linewidth]{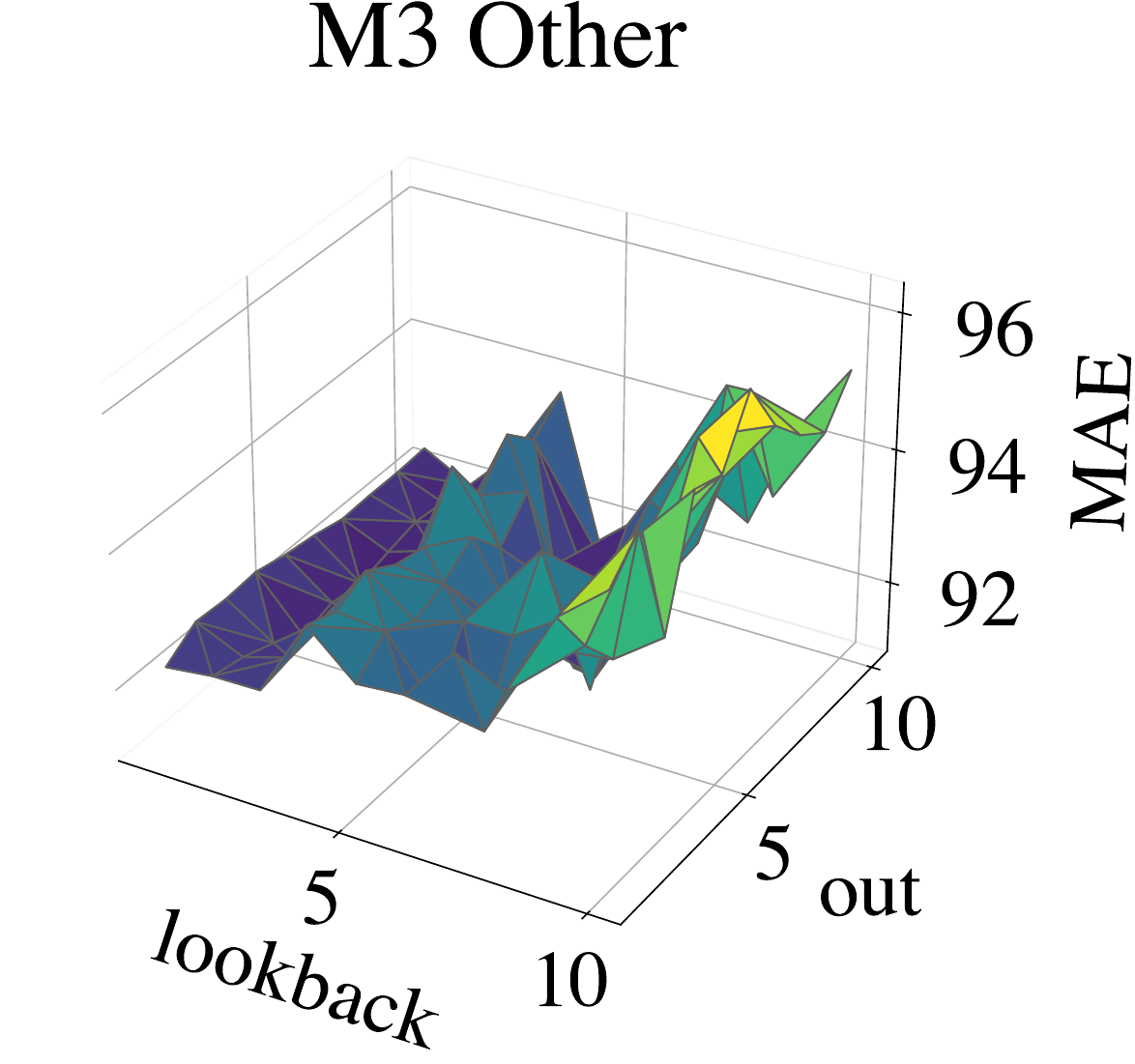}
	}\\
	\caption{Relationship between  $error$, $lookback$ and $out$ in M3 data set}
\end{figure*}

\textit{DFCNN} has 2 hyperparameter windows size $lookback$ and quantity of convolution kernels $out$. Window size $lookback$ determines the length of fuzzy rules. Traditional \textit{FTSF} is only able to generate the fuzzy rules with length 2. \textit{DFCNN} utilizes Sliding Windows algorithm to expand the fuzzy rules' length. Different fuzzy rules' length contains different local information and observation field. The longer windows size $lookback$ is, the more local information is embedded. The quantity of convolution kernels $out$  is also called the output channels of convolution layer. Convolution kernel is mutually independent. The more convolution kernels work together, the more output channels generated which increases the possibility of better regression. But it is not absolute that forecasting error in longer $lookback$ and more $out$ is lower than one in shorter $lookback$ or less $out$.

For $lookback$ and $out$ in parameter setting, the default $lookback=2$ setting refers to length of fuzzy rules in \textit{FTSF}, while the out parameter is set as $out=2$ in order to improve the stability of the method instead of $out=1$.  \textit{DFCNN} conducted a total of $100$ parameter portfolio experiments in both $lookback \in [1,10]$ and $out \in [1,10]$. 

The experimental results are presented in Fig.10 with a $lookback-out-error$ three-dimensional coordinate surface. The fact is that the forecasting error varies more with window size $lookback$ than out channels $out$. When the window size $lookback$ is specified, the forecasting error has limited variation with the output channels $out$. The prime way to improve the forecasting accuracy of \textit{DFCNN} in the practice is to adjust the windows size $lookback$. Adjusting the out channels is suitable for further improving accuracy after adjusting the windows size $lookback$.

For a certain time series, the best hyperparameters is not absolute. In the practice, the windows size $lookback$ is recommended to set $lookback=2$ which follow the traditional \textit{FTSF}. Windows size $lookback$ increasing involves more local information but there is possible to mislead \textit{DFCNN} to recognize meaningless feature and decrease the forecasting accuracy. The best practice is increasing windows size $lookback$ step by step. In M3 Quarterly, \textit{DFCNN} obtains the best performance with $lookback=4$. Then considering adjust out channels is a useful way of subtle enhancements. A word of warning is that too many output channels can lead to overfitting problems, which is a common problem with neural networks. In the M3 Other dataset, when $lookback=3$, the increase in the number of convolutional kernels approaches $out=10$, and the error increases sharply, which is a sign of overfitting.

\subsection{Discussion And Future Direction}

\textit{DFCNN} is an improved FTSF method combined neural network. \textit{DFCNN} absorbs the advantages of the previous work: considering the time series trend, using the \textbf{difference algorithm} to remove the non-stationarity, rather than weighting the trend;\textbf{ the length of fuzzy relation is expanded} instead of the input of fuzzy relation; \textbf{the model optimizes the weights globally} using learning algorithms, rather than learning only in fuzzy reasoning.

\textit{DFCNN} aims at the goal of \textbf{single-step time series forecasting with single variable}, especially for time series with trends such as stock forecasting. \textit{DFCNN} can effectively deal with the trend of stocks. Due to the existence of the difference algorithm, each data is dependent on the previous time step. So it is important to note that \textit{DFCNN} cannot predict multiple steps at once. Therefore, it is suitable to predict time series with time intervals, such as the number of school enrollments, the number of infections or the national GDP. \textbf{According to the experimental results}, \textit{DFCNN} has low time complexity compared with the existing models, and can complete the task of low forecasting error when \textbf{facing different types of time series}. Another advantage of deploying \textit{DFCNN} is that the \textit{DFCNN} model is lightweight, does not have the data requirements such as large models, and the training period is short. \textit{DFCNN} is suitable for data scientists to analyze and evaluate time series. 

\textbf{For multi-step forecasting}, \textit{DFCNN} can only be combined by multiple single-step forecastings, with the last forecasting result as the input. That will seriously reduce the forecasting accuracy, so \textit{DFCNN} needs to be further improved to be compatible with multi-step forecasting. At this point, the difference algorithm needs to be replaced to remove the correlation between the data, while ensuring that the trend information is processed efficiently. In terms of periodicity, DFCNN does not carry out too much processing, because the window size parameter includes periodic information to a certain extent. \textbf{In terms of periodic model improvement}, the structure of convolutional layer and input fuzzy information can be improved by referring to RNN model structure. However, it should be noted that the parameter size of the model is too large, resulting in incomplete training. \textbf{In terms of the number of variables or input channels}, most \textit{FTSF} work belong to univariate model. Because \textit{Fuzzy Token} is proposed, which belongs to the representation method of fuzzy time series, the multi-channel learning model can be expanded based on \textit{Fuzzy Token}. \textit{Fuzzy Token} is a vector that can be optimized again for compatibility with an existing multivariate backone.
 
\section{Conclusion}
In this article, a imporved \textit{FTSF} model \textit{DFCNN} is proposed. Primarily, the process of \textit{FTSF} is analyzed and limitation of \textit{FTSF} is idea of expert system.  \textit{Fuzzy Generator} represents Fuzzy time series in the form of vector \textit{Fuzzy Token}. Hence \textit{Convolution Layer} achieves equivalent fuzzification and implementation of learnable properties. Then, \textit{DFCNN} considers the stationary of time series and solve the FF problem of traditional \textit{FTSF} by Difference algorithm. Besides, \textit{DFCNN} further expands the length of fuzzy rules to provide more possibility of learning local information. Expanding the length of fuzzy rules also provides more chances to improve forecasting accuracy. 

In the experimental part, the \textit{DFCNN} was tested on the CCI and M competition datasets. Compared with the traditional \textit{FTSF}, DFCNN absorbs the strengths of previous \textit{FTSF} work. Compared with common time series forecasting methods, \textit{DFCNN} still leads most of them. Under the verification of different data amounts and different types of time series, \textit{DFCNN} has the stability similar to statistical models, and has the learning ability similar to machine learning and deep learning models, which further reduces the error.

\textit{DFCNN} is suitable for application in single variable single step time series tasks. \textit{DFCNN} is particularly suitable for data with significant trends such as national GDP, population and stock forecasts. The lightweight construction also makes the \textit{DFCNN} easy to train and adjust parameters. In the future, the combination of \textit{FTSF} and deep learning needs to focus on how to \textbf{break through the limit of single step forecasting} and how to adapt to \textbf{multi-variable time series forecasting}, which will \textbf{further broaden the application scenario of \textit{FTSF}}.


\bibliographystyle{IEEEtran}
\bibliography{reference}

\begin{thebibliography}{10}
\providecommand{\url}[1]{#1}
\csname url@samestyle\endcsname
\providecommand{\newblock}{\relax}
\providecommand{\bibinfo}[2]{#2}
\providecommand{\BIBentrySTDinterwordspacing}{\spaceskip=0pt\relax}
\providecommand{\BIBentryALTinterwordstretchfactor}{4}
\providecommand{\BIBentryALTinterwordspacing}{\spaceskip=\fontdimen2\font plus
\BIBentryALTinterwordstretchfactor\fontdimen3\font minus
  \fontdimen4\font\relax}
\providecommand{\BIBforeignlanguage}[2]{{%
\expandafter\ifx\csname l@#1\endcsname\relax
\typeout{** WARNING: IEEEtran.bst: No hyphenation pattern has been}%
\typeout{** loaded for the language `#1'. Using the pattern for}%
\typeout{** the default language instead.}%
\else
\language=\csname l@#1\endcsname
\fi
#2}}
\providecommand{\BIBdecl}{\relax}
\BIBdecl

\bibitem{blazquez2021water}
A.~Bl{\'a}zquez-Garc{\'\i}a, A.~Conde, U.~Mori, and J.~A. Lozano, ``Water leak
  detection using self-supervised time series classification,''
  \emph{Information Sciences}, vol. 574, pp. 528--541, 2021.

\bibitem{wang2022time}
W.~Wang, W.~Liu, and H.~Chen, ``Time series forecasting via fuzzy-probabilistic
  approach with evolving clustering-based granulation,'' \emph{IEEE
  Transactions on Fuzzy Systems}, 2022.

\bibitem{hayashi2022ocstn}
T.~Hayashi, D.~Cimr, F.~Studni{\v{c}}ka, H.~Fujita, D.~Bu{\v{s}}ovsk{\`y}, and
  R.~Cimler, ``Ocstn: One-class time-series classification approach using a
  signal transformation network into a goal signal,'' \emph{Information
  Sciences}, vol. 614, pp. 71--86, 2022.

\bibitem{liu2022time}
S.~Liu, B.~Zhou, Q.~Ding, B.~Hooi, Z.~bo~Zhang, H.~Shen, and X.~Cheng, ``Time
  series anomaly detection with adversarial reconstruction networks,''
  \emph{IEEE Transactions on Knowledge and Data Engineering}, 2022.

\bibitem{karczmarek2022choquet}
P.~Karczmarek, {\L}.~Ga{\l}ka, A.~Kiersztyn, M.~Dolecki, K.~Kiersztyn, and
  W.~Pedrycz, ``Choquet integral-based aggregation for the analysis of
  anomalies occurrence in sustainable transportation systems,'' \emph{IEEE
  Transactions on Fuzzy Systems}, 2022.

\bibitem{peng2022characterizing}
K.~Peng and P.~Shang, ``Characterizing ordinal network of time series based on
  complexity-entropy curve,'' \emph{Pattern Recognition}, vol. 124, p. 108464,
  2022.

\bibitem{cui2022belief}
H.~Cui, L.~Zhou, Y.~Li, and B.~Kang, ``Belief entropy-of-entropy and its
  application in the cardiac interbeat interval time series analysis,''
  \emph{Chaos, Solitons \& Fractals}, vol. 155, p. 111736, 2022.

\bibitem{hernandez2020forecasting}
A.~Hernandez-Matamoros, H.~Fujita, T.~Hayashi, and H.~Perez-Meana,
  ``Forecasting of covid19 per regions using arima models and polynomial
  functions,'' \emph{Applied soft computing}, vol.~96, p. 106610, 2020.

\bibitem{wang2022trend}
Y.~Wang, F.~Yu, W.~Homenda, W.~Pedrycz, Y.~Tang, A.~Jastrzebska, and F.~Li,
  ``The trend-fuzzy-granulation-based adaptive fuzzy cognitive map for
  long-term time series forecasting,'' \emph{IEEE Transactions on Fuzzy
  Systems}, 2022.

\bibitem{zhang2022interpretable}
D.~Zhang, Y.~Xu, Y.~Peng, C.~Du, N.~Wang, M.~Tang, L.~Lu, and J.~Liu, ``An
  interpretable station delay prediction model based on graph community neural
  network and time-series fuzzy decision tree,'' \emph{IEEE Transactions on
  Fuzzy Systems}, 2022.

\bibitem{huang2021natural}
Y.~Huang, X.~Mao, and Y.~Deng, ``Natural visibility encoding for time series
  and its application in stock trend prediction,'' \emph{Knowledge-Based
  Systems}, vol. 232, p. 107478, 2021.

\bibitem{billah2006exponential}
B.~Billah, M.~L. King, R.~D. Snyder, and A.~B. Koehler, ``Exponential smoothing
  model selection for forecasting,'' \emph{International journal of
  forecasting}, vol.~22, no.~2, pp. 239--247, 2006.

\bibitem{jiang2020holt}
W.~Jiang, X.~Wu, Y.~Gong, W.~Yu, and X.~Zhong, ``Holt--winters smoothing
  enhanced by fruit fly optimization algorithm to forecast monthly electricity
  consumption,'' \emph{Energy}, vol. 193, p. 116779, 2020.

\bibitem{xu2019modeling}
H.~Xu, F.~Ding, and E.~Yang, ``Modeling a nonlinear process using the
  exponential autoregressive time series model,'' \emph{Nonlinear Dynamics},
  vol.~95, no.~3, pp. 2079--2092, 2019.

\bibitem{domingos2019intelligent}
S.~d.~O. Domingos, J.~F. de~Oliveira, and P.~S. de~Mattos~Neto, ``An
  intelligent hybridization of arima with machine learning models for time
  series forecasting,'' \emph{Knowledge-Based Systems}, vol. 175, pp. 72--86,
  2019.

\bibitem{liu2021short}
X.~Liu, Z.~Lin, and Z.~Feng, ``Short-term offshore wind speed forecast by
  seasonal arima-a comparison against gru and lstm,'' \emph{Energy}, vol. 227,
  p. 120492, 2021.

\bibitem{masini2021machine}
R.~P. Masini, M.~C. Medeiros, and E.~F. Mendes, ``Machine learning advances for
  time series forecasting,'' \emph{Journal of Economic Surveys}, 2021.

\bibitem{huang2021new}
Y.~Huang and Y.~Deng, ``A new crude oil price forecasting model based on
  variational mode decomposition,'' \emph{Knowledge-Based Systems}, vol. 213,
  p. 106669, 2021.

\bibitem{ponce2020tailored}
P.~Ponce, A.~Meier, J.~I. M{\'e}ndez, T.~Peffer, A.~Molina, and O.~Mata,
  ``Tailored gamification and serious game framework based on fuzzy logic for
  saving energy in connected thermostats,'' \emph{Journal of Cleaner
  Production}, vol. 262, p. 121167, 2020.

\bibitem{wu2021strategies}
Y.~Wu, B.~Kang, and H.~Wu, ``Strategies of attack--defense game for wireless
  sensor networks considering the effect of confidence level in fuzzy
  environment,'' \emph{Engineering Applications of Artificial Intelligence},
  vol. 102, p. 104238, 2021.

\bibitem{xu2022game}
X.~Xu, Q.~Jiang, P.~Zhang, X.~Cao, M.~R. Khosravi, L.~T. Alex, L.~Qi, and
  W.~Dou, ``Game theory for distributed iov task offloading with fuzzy neural
  network in edge computing,'' \emph{IEEE Transactions on Fuzzy Systems}, 2022.

\bibitem{Xiao2021GIQ}
F.~Xiao, ``{GIQ: A generalized intelligent quality-based approach for fusing
  multi-source information},'' \emph{IEEE Transactions on Fuzzy Systems},
  vol.~29, no.~7, pp. 2018--2031, 2021.

\bibitem{albahri2022novel}
O.~Albahri, A.~Zaidan, A.~Albahri, H.~Alsattar, R.~Mohammed, U.~Aickelin,
  G.~Kou, F.~Jumaah, M.~M. Salih, A.~Alamoodi \emph{et~al.}, ``Novel dynamic
  fuzzy decision-making framework for covid-19 vaccine dose recipients,''
  \emph{Journal of advanced research}, vol.~37, pp. 147--168, 2022.

\bibitem{xiao2020efmcdm}
F.~Xiao, ``{EFMCDM: Evidential fuzzy multicriteria decision making based on
  belief entropy},'' \emph{IEEE Transactions on Fuzzy Systems}, vol.~28, no.~7,
  pp. 1477--1491, 2020.

\bibitem{kumar2022multiple}
K.~Kumar and S.-M. Chen, ``Multiple attribute group decision making based on
  advanced linguistic intuitionistic fuzzy weighted averaging aggregation
  operator of linguistic intuitionistic fuzzy numbers,'' \emph{Information
  Sciences}, vol. 587, pp. 813--824, 2022.

\bibitem{Xiao2022Acomplexweighted}
F.~Xiao, Z.~Cao, and C.-T. Lin, ``A complex weighted discounting multisource
  information fusion with its application in pattern classification,''
  \emph{IEEE Transactions on Knowledge and Data Engineering}, p. DOI:
  10.1109/TKDE.2022.3206871, 2022.

\bibitem{Xiao2022GEJS}
F.~Xiao, ``{GEJS: A generalized evidential divergence measure for multisource
  information fusion},'' \emph{IEEE Transactions on Systems, Man, and
  Cybernetics - Systems}, p. DOI: 10.1109/TSMC.2022.3211498, 2022.

\bibitem{zhang2020boosting}
J.~Zhang, J.~X. Huang, and Q.~V. Hu, ``Boosting evolutionary optimization via
  fuzzy-classification-assisted selection,'' \emph{Information Sciences}, vol.
  519, pp. 423--438, 2020.

\bibitem{zhang2021lfic}
H.~Zhang, S.~Zhong, Y.~Deng, and K.~H. Cheong, ``Lfic: Identifying influential
  nodes in complex networks by local fuzzy information centrality,'' \emph{IEEE
  Transactions on Fuzzy Systems}, 2021.

\bibitem{xiao2021distance}
F.~Xiao, ``A distance measure for intuitionistic fuzzy sets and its application
  to pattern classification problems,'' \emph{IEEE Transactions on Systems,
  Man, and Cybernetics: Systems}, vol.~51, no.~6, pp. 3980--3992, 2021.

\bibitem{rashid2022novel}
J.~Rashid, J.~Kim, A.~Hussain, U.~Naseem, and S.~Juneja, ``A novel multiple
  kernel fuzzy topic modeling technique for biomedical data,'' \emph{BMC
  bioinformatics}, vol.~23, no.~1, pp. 1--19, 2022.

\bibitem{barua2023automated}
P.~D. Barua, T.~Keles, S.~Dogan, M.~Baygin, T.~Tuncer, C.~F. Demir, H.~Fujita,
  R.-S. Tan, C.~P. Ooi, and U.~R. Acharya, ``Automated eeg sentence
  classification using novel dynamic-sized binary pattern and multilevel
  discrete wavelet transform techniques with tseeg database,'' \emph{Biomedical
  Signal Processing and Control}, vol.~79, p. 104055, 2023.

\bibitem{Xiao2022NQMF}
F.~Xiao and W.~Pedrycz, ``Negation of the quantum mass function for multisource
  quantum information fusion with its application to pattern classification,''
  \emph{IEEE Transactions on Pattern Analysis and Machine Intelligence}, p.
  DOI: 10.1109/TPAMI.2022.3167045, 2022.

\bibitem{Xiao2022Generalizeddivergence}
F.~Xiao, J.~Wen, and W.~Pedrycz, ``Generalized divergence-based decision making
  method with an application to pattern classification,'' \emph{IEEE
  Transactions on Knowledge and Data Engineering}, p. DOI:
  10.1109/TKDE.2022.3177896, 2022.

\bibitem{cheng2022ranking}
R.~Cheng, J.~Zhang, and B.~Kang, ``Ranking of z-numbers based on the developed
  golden rule representative value,'' \emph{IEEE Transactions on Fuzzy
  Systems}, vol.~30, no.~12, pp. 5196--5210, 2022.

\bibitem{lin2022picture}
M.~Lin, X.~Li, R.~Chen, H.~Fujita, and J.~Lin, ``Picture fuzzy interactional
  partitioned heronian mean aggregation operators: an application to madm
  process,'' \emph{Artificial Intelligence Review}, vol.~55, no.~2, pp.
  1171--1208, 2022.

\bibitem{cleveland1990stl}
R.~B. Cleveland, W.~S. Cleveland, J.~E. McRae, and I.~Terpenning, ``Stl: A
  seasonal-trend decomposition,'' \emph{J. Off. Stat}, vol.~6, no.~1, pp.
  3--73, 1990.

\bibitem{rahman2015artificial}
N.~H.~A. Rahman, M.~H. Lee, and M.~T. Latif, ``Artificial neural networks and
  fuzzy time series forecasting: an application to air quality,'' \emph{Quality
  \& Quantity}, vol.~49, pp. 2633--2647, 2015.

\bibitem{singh2018rainfall}
P.~Singh, ``Rainfall and financial forecasting using fuzzy time series and
  neural networks based model,'' \emph{International Journal of Machine
  Learning and Cybernetics}, vol.~9, no.~3, pp. 491--506, 2018.

\bibitem{jiang2019hybrid}
P.~Jiang, H.~Yang, and J.~Heng, ``A hybrid forecasting system based on fuzzy
  time series and multi-objective optimization for wind speed forecasting,''
  \emph{Applied energy}, vol. 235, pp. 786--801, 2019.

\bibitem{kocak2020new}
C.~Kocak, A.~Z. Dalar, O.~Cagcag~Yolcu, E.~Bas, and E.~Egrioglu, ``A new fuzzy
  time series method based on an arma-type recurrent pi-sigma artificial neural
  network,'' \emph{Soft Computing}, vol.~24, pp. 8243--8252, 2020.

\bibitem{bas2022novel}
E.~Bas, E.~Egrioglu, and E.~Kolemen, ``A novel intuitionistic fuzzy time series
  method based on bootstrapped combined pi-sigma artificial neural network,''
  \emph{Engineering Applications of Artificial Intelligence}, vol. 114, p.
  105030, 2022.

\bibitem{egrioglu_new_2009}
E.~Egrioglu, C.~H. Aladag, U.~Yolcu, V.~R. Uslu, and M.~A. Basaran,
  ``\BIBforeignlanguage{en-US}{A new approach based on artificial neural
  networks for high order multivariate fuzzy time series},''
  \emph{\BIBforeignlanguage{en-US}{Expert Systems with Applications}}, vol.~36,
  no.~7, pp. 10\,589--10\,594, Sep. 2009, wOS:000266851000044.

\bibitem{aladag2009forecasting}
C.~H. Aladag, M.~A. Basaran, E.~Egrioglu, U.~Yolcu, and V.~R. Uslu,
  ``Forecasting in high order fuzzy times series by using neural networks to
  define fuzzy relations,'' \emph{Expert Systems with Applications}, vol.~36,
  no.~3, pp. 4228--4231, 2009.

\bibitem{aladag2010high}
C.~H. Aladag, U.~Yolcu, and E.~Egrioglu, ``A high order fuzzy time series
  forecasting model based on adaptive expectation and artificial neural
  networks,'' \emph{Mathematics and Computers in Simulation}, vol.~81, no.~4,
  pp. 875--882, 2010.

\bibitem{chen2011multivariate}
S.-M. Chen and K.~Tanuwijaya, ``Multivariate fuzzy forecasting based on fuzzy
  time series and automatic clustering techniques,'' \emph{Expert Systems with
  Applications}, vol.~38, no.~8, pp. 10\,594--10\,605, 2011.

\bibitem{egrioglu2013fuzzy}
E.~Egrioglu, C.~H. Aladag, and U.~Yolcu, ``Fuzzy time series forecasting with a
  novel hybrid approach combining fuzzy c-means and neural networks,''
  \emph{Expert Systems with Applications}, vol.~40, no.~3, pp. 854--857, 2013.

\bibitem{chen1996forecasting}
S.-M. Chen, ``Forecasting enrollments based on fuzzy time series,'' \emph{Fuzzy
  sets and systems}, vol.~81, no.~3, pp. 311--319, 1996.

\bibitem{Goodfellow-et-al-2016}
I.~Goodfellow, Y.~Bengio, and A.~Courville, \emph{Deep Learning}.\hskip 1em
  plus 0.5em minus 0.4em\relax MIT Press, 2016,
  \url{http://www.deeplearningbook.org}.

\bibitem{song2022dynamic}
J.~Song, Y.-K. Wang, and Y.~Niu, ``Dynamic event-triggered terminal sliding
  mode control under binary encoding: Analysis and experimental validation,''
  \emph{IEEE Transactions on Circuits and Systems I: Regular Papers}, vol.~69,
  no.~9, pp. 3772--3782, 2022.

\bibitem{ioffe2015batch}
S.~Ioffe and C.~Szegedy, ``Batch normalization: Accelerating deep network
  training by reducing internal covariate shift,'' in \emph{International
  conference on machine learning}.\hskip 1em plus 0.5em minus 0.4em\relax PMLR,
  2015, pp. 448--456.

\bibitem{awais2020revisiting}
M.~Awais, M.~T.~B. Iqbal, and S.-H. Bae, ``Revisiting internal covariate shift
  for batch normalization,'' \emph{IEEE Transactions on Neural Networks and
  Learning Systems}, vol.~32, no.~11, pp. 5082--5092, 2020.

\bibitem{ashuri2010time}
B.~Ashuri and J.~Lu, ``Time series analysis of enr construction cost index,''
  \emph{Journal of Construction Engineering and Management}, vol. 136, no.~11,
  pp. 1227--1237, 2010.

\bibitem{godahewa2021monash}
R.~Godahewa, C.~Bergmeir, G.~I. Webb, R.~J. Hyndman, and P.~Montero-Manso,
  ``Monash time series forecasting archive,'' \emph{arXiv preprint
  arXiv:2105.06643}, 2021.

\bibitem{cheng2009forecasting}
C.-H. Cheng, Y.-S. Chen, and Y.-L. Wu, ``Forecasting innovation diffusion of
  products using trend-weighted fuzzy time-series model,'' \emph{Expert Systems
  with Applications}, vol.~36, no.~2, pp. 1826--1832, 2009.

\bibitem{severiano2017very}
C.~A. Severiano, P.~C. Silva, H.~J. Sadaei, and F.~G. Guimar{\~a}es, ``Very
  short-term solar forecasting using fuzzy time series,'' in \emph{2017 IEEE
  international conference on fuzzy systems (FUZZ-IEEE)}.\hskip 1em plus 0.5em
  minus 0.4em\relax IEEE, 2017, pp. 1--6.

\bibitem{hyndman2008forecasting}
R.~Hyndman, A.~B. Koehler, J.~K. Ord, and R.~D. Snyder, \emph{Forecasting with
  exponential smoothing: the state space approach}.\hskip 1em plus 0.5em minus
  0.4em\relax Springer Science \& Business Media, 2008.

\bibitem{hyndman2018forecasting}
R.~J. Hyndman and G.~Athanasopoulos, \emph{Forecasting: principles and
  practice}.\hskip 1em plus 0.5em minus 0.4em\relax OTexts, 2018.

\bibitem{trapero2015identification}
J.~R. Trapero, N.~Kourentzes, and R.~Fildes, ``On the identification of sales
  forecasting models in the presence of promotions,'' \emph{Journal of the
  operational Research Society}, vol.~66, pp. 299--307, 2015.

\bibitem{prokhorenkova2018catboost}
L.~Prokhorenkova, G.~Gusev, A.~Vorobev, A.~V. Dorogush, and A.~Gulin,
  ``Catboost: unbiased boosting with categorical features,'' \emph{Advances in
  neural information processing systems}, vol.~31, 2018.

\bibitem{salinas2020deepar}
D.~Salinas, V.~Flunkert, J.~Gasthaus, and T.~Januschowski, ``Deepar:
  Probabilistic forecasting with autoregressive recurrent networks,''
  \emph{International Journal of Forecasting}, vol.~36, no.~3, pp. 1181--1191,
  2020.

\bibitem{borovykh2017conditional}
A.~Borovykh, S.~Bohte, and C.~W. Oosterlee, ``Conditional time series
  forecasting with convolutional neural networks,'' \emph{arXiv preprint
  arXiv:1703.04691}, 2017.

\bibitem{vaswani2017attention}
A.~Vaswani, N.~Shazeer, N.~Parmar, J.~Uszkoreit, L.~Jones, A.~N. Gomez,
  {\L}.~Kaiser, and I.~Polosukhin, ``Attention is all you need,''
  \emph{Advances in neural information processing systems}, vol.~30, 2017.

\end{thebibliography}

\newpage

\end{document}